\documentclass[10pt,twocolumn,letterpaper]{article}

\usepackage[pagenumbers]{cvpr} %
\newif\ifarxiv
\arxivtrue %

\usepackage[dvipsnames]{xcolor}

\usepackage{times}
\usepackage{microtype}
\usepackage{epsfig}
\usepackage[table,xcdraw,dvipsnames]{xcolor}
\usepackage[skip=5pt]{caption}
\usepackage{float}
\usepackage{placeins}
\usepackage{color, colortbl}
\usepackage{stfloats}
\usepackage{enumitem}
\usepackage{tabularx}
\usepackage{xstring}
\usepackage{cuted} %
\usepackage{multirow}
\usepackage{xspace}
\usepackage{url}
\usepackage{subcaption}
\usepackage{mathtools}
\usepackage[hang,flushmargin]{footmisc}
\usepackage[normalem]{ulem}  %
\usepackage{graphicx}
\usepackage{amsmath}
\usepackage{amssymb}
\usepackage{comment}
\usepackage[title]{appendix}
\usepackage{cancel}
\usepackage{bbm}
\usepackage{afterpage}
\usepackage{algorithm}
\usepackage{algorithmic}
\usepackage{bm}
\usepackage{gensymb}
\usepackage{tikz}
\usepackage{pifont}

\newfloat{sepfigure}{p}{lop}
\floatname{sepfigure}{Figure}

\definecolor{R1color}{rgb}{0,0.56,0}
\newcommand{\R}[1]{{%
    \textbf{%
        \ifstrequal{#1}{1}{\textcolor[rgb]{0,0.56,0}{R#1}}{%
        \ifstrequal{#1}{2}{\textcolor{blue}{R#1}}{%
        \ifstrequal{#1}{3}{\textcolor{magenta}{R#1}}{%
        \ifstrequal{#1}{4}{\textcolor{teal}{R#1}}{%
                           \textcolor{cyan}{R#1}%
        }}}}%
    }%
}}

\usepackage{algorithm}
\usepackage{algorithmic}

\usepackage{bm}

\usepackage{comment}
\usepackage{float}

\def\paperID{5283}
\def\confName{CVPR}
\def\confYear{2026}
\def\paperTitle{Dexterous World Models}

\def\authorBlock{
    Byungjun Kim$^{1}$\thanks{Equal contribution} \qquad
    Taeksoo Kim$^{1}$\footnotemark[1] \qquad
    Junyoung Lee$^{1}$ \qquad
    Hanbyul Joo$^{1,2}$ \\
    [0.4em] %
    $^1$Seoul National University \quad $^2$RLWRLD \\
    {\tt\small \href{https://snuvclab.github.io/dwm/}{\color{magenta}{https://snuvclab.github.io/dwm/}}}
}

\definecolor{cvprblue}{rgb}{0.21,0.49,0.74}
\usepackage[pagebackref,breaklinks,colorlinks,citecolor=cvprblue]{hyperref}

\title{\paperTitle}

\author{\authorBlock}

\begin{document}
\maketitle

\begin{strip}
    \vspace{-15pt}
    \centering
    \begin{minipage}{\textwidth}
        \centering
        \includegraphics[width=0.9\linewidth, trim={0cm 0cm 0cm 0cm}]{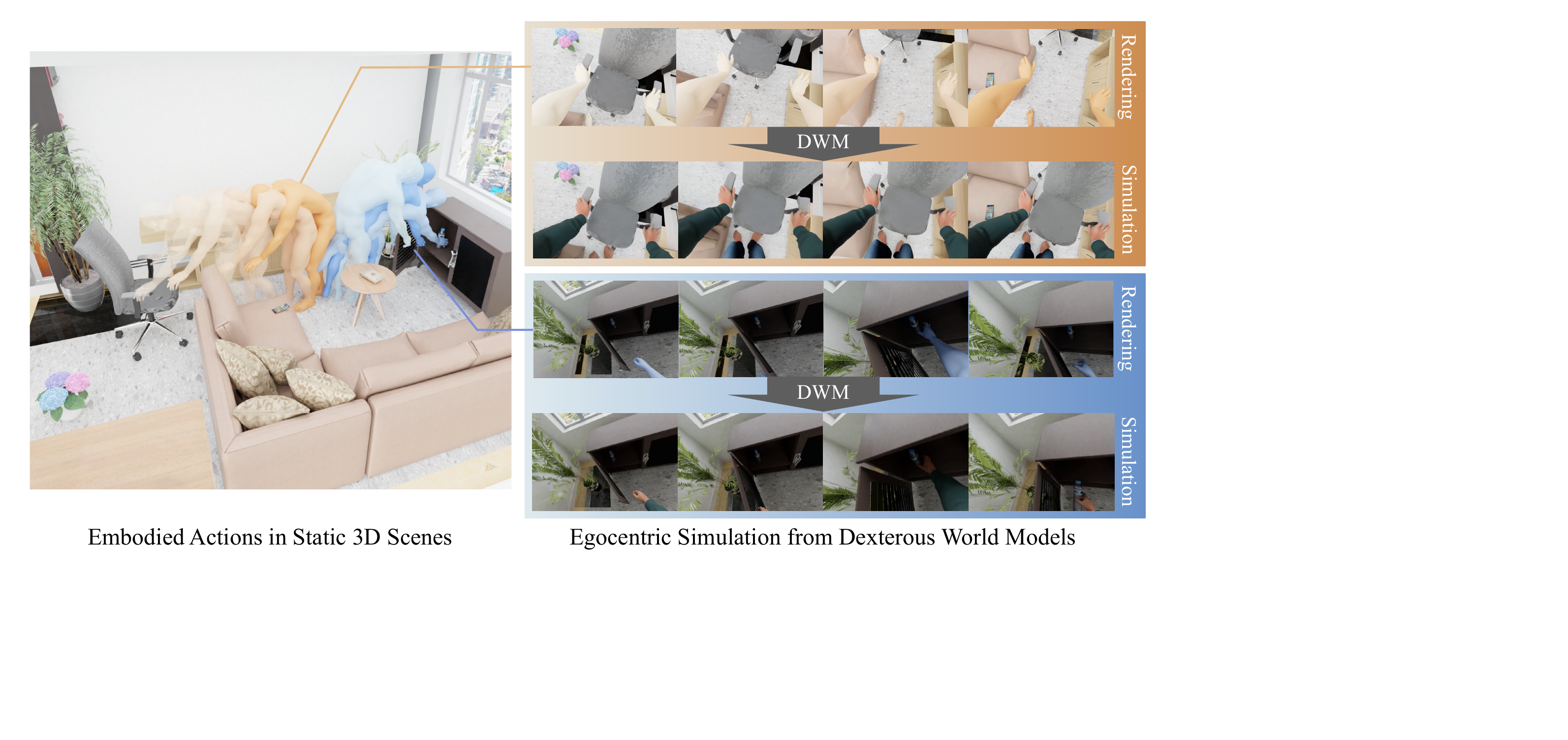}
        \captionof{figure}{
            \textbf{Dexterous World Models} predict egocentric visual dynamics of static 3D scenes, driven by dexterous hand manipulations.
            }
        \label{fig:teaser}
    \end{minipage}
\end{strip}

\begin{abstract}
Recent progress in 3D reconstruction has made it easy to create realistic digital twins from everyday environments. However, current digital twins remain largely static—limited to navigation and view synthesis without embodied interactivity. To bridge this gap, we introduce Dexterous World Model (DWM), a scene-action-conditioned video diffusion framework that models how dexterous human actions induce dynamic changes in static 3D scenes. 
Given a static 3D scene rendering and an egocentric hand motion sequence, DWM generates temporally coherent videos depicting plausible human–scene interactions. Our approach conditions video generation on (1) static scene renderings following a specified camera trajectory to ensure spatial consistency, and (2) egocentric hand mesh renderings that encode both geometry and motion cues in the egocentric view to model action-conditioned dynamics directly. To train DWM, we construct a hybrid interaction video dataset: synthetic egocentric interactions provide fully aligned supervision for joint locomotion–manipulation learning, while fixed-camera real-world videos contribute diverse and realistic object dynamics.
Experiments demonstrate that DWM enables realistic, physically plausible interactions, such as grasping, opening, or moving objects, while maintaining camera and scene consistency. This framework establishes the first step toward video diffusion-based interactive digital twins, enabling embodied simulation from egocentric actions.
\end{abstract}
    
\section{Introduction} \label{sec:intro} 

World models~\cite{ha2018worldmodel} aim to capture the underlying structure and dynamics of the environment so that intelligent systems can reason, plan, and act effectively. Modeling dynamic consequences and causal relationships is central to world modeling, which enables the prediction of environmental changes resulting from a wide range of actions. Notably, many of the most important and complex dynamics are driven by human-environment interactions, where humans manipulate objects and actively alter the state of the world.
A practical world model must therefore: (1) maintain a consistent representation of the static components of a scene, (2) accept an action specification (e.g., dexterous hand manipulation) that induces dynamic changes, and (3) generate the resulting dynamics while faithfully preserving the unaltered regions.

While video generative models~\cite{yang2025cogvideox} provide a promising foundation for world modeling, they remain insufficient to function as a faithful world model. These models typically attempt to synthesize the entire scene, both static and dynamic regions, making it challenging to focus on the specific dynamic changes induced by action inputs. Furthermore, most video models accept only textual instruction as ``action input". However, textual descriptions are inherently imprecise, as they cannot accurately specify human actions such as hand configurations, fine-grained temporal control, both of which are crucial for modeling realistic human manipulations. 
Many existing world models often treat camera motion as the action input~\cite{zhu2025aether, bar2025nwm}. However, camera motion is not the primary drivers of world dynamics. The most significant source of changes in everyday settings is dexterous hand interaction, such as grasping and object manipulation, which are absent from existing approaches. 

In this paper, we present \textbf{Dexterous World Models (DWM)}, a new formulation of world modeling that takes both a static scene and dexterous human hand motions as input, and predicts the dynamic changes induced by those actions (\cref{fig:teaser}).
Unlike existing approaches that require the model to synthesize the entire scene, both static and dynamic, we explicitly provide a rendering of the static environment as input, together with dexterous hand trajectories captured from an egocentric perspective. This design reflects a more natural setting because humans also perceive and internalize the static world before acting upon it, encouraging the model to focus on the dynamic changes of the world induced by human actions, while preserving others. 
The egocentric formulation further provides inherent grounding for interaction because it naturally captures the user’s focus of attention and hand trajectory during manipulation. 

To realize our new formulation, we implement DWM as a \emph{scene-action-conditioned video diffusion model} that enables action-conditioned simulation in static 3D scenes. DWM is designed to capture how human actions induce dynamic changes in an otherwise static environment, directly modeling realistic scene interactions driven by dexterous hand actions.
Our approach conditions the diffusion process on two inputs: (1) a static scene video rendered along a specified camera trajectory to ensure spatial consistency, and (2) egocentric hand mesh renderings that encode geometric and motion cues for modeling action-conditioned dynamics.
A core idea behind our formulation is to model only the residual dynamics caused by human actions while preserving all unaltered regions.
To achieve this, we initialize DWM from a pretrained video inpainting diffusion model~\cite{VideoX-Fun}. When given a full mask, such models reproduce the input video, effectively behaving as an identity mapping with a generative prior.
Under this initialization, using the static scene video as input establishes a base output that preserves the egocentric camera motion and scene appearance without introducing interaction dynamics.
Conditioning the model on dexterous egocentric hand motions then guides the diffusion process to synthesize only manipulation-induced changes.

As no existing dataset provides paired static scene renderings and corresponding interaction videos captured under identical camera trajectories required by our formulation, we leverage a synthetic 3D human–scene interaction dataset~\cite{jiang2024trumans} to construct aligned interaction videos, static scene videos, and hand videos.
While synthetic data provides precise alignment under identical camera trajectories, it offers limited diversity in interaction dynamics. We therefore additionally incorporate fixed-camera real-world interaction videos~\cite{zhao2025taste}, enriching the model with diverse and realistic interaction dynamics.
Our experiments show that DWM generalizes to real-world scenarios involving combined locomotion and manipulation.
Beyond video synthesis, we demonstrate that DWM can be used as a visual world model for simulation-based action evaluation by comparing the predicted visual outcomes of candidate actions against a given goal.

Our main contributions can be summarized as follows:
(1) We introduce Dexterous World Models (DWM), a new formulation of world modeling via scene-action-conditioned video diffusion.
(2) We propose a conditioning scheme that leverages static scene videos and egocentric hand motions to model interaction-induced residual dynamics.
(3) We construct a hybrid training dataset combining synthetic egocentric interactions with fixed-camera real-world interaction videos, which provides aligned supervision and realistic interaction dynamics.
(4) We show that DWM can be used for simulation-based reasoning by simulating the visual consequences of candidate actions.

\section{Related Work}
\label{sec:related}

\paragraph{World Models.}
\label{sec:rel_world_models}
World models aim to learn predictive environment dynamics, enabling agents to anticipate how the world evolves in response to actions. 
The classical formulation of \citet{ha2018worldmodel} demonstrated that compact latent dynamics can support imagination-based planning, inspiring extensive work that integrates learned world models with reinforcement learning for control and decision-making~\cite{hafner2019planet,hafner2025dreamerv3,zhou2024dinowmworldmodelspretrained}. 
While effective, these approaches typically operate in low-dimensional latent spaces and do not model rich visual interactions.
To capture full visual scene evolution beyond latent dynamics, recent work adopts video generative models to build visual world models that predict future observations directly in pixel space. 
Action conditioning in these models commonly relies on camera pose changes~\cite{bar2025nwm} or Plücker-ray encodings~\cite{zhu2025aether,zhou2025seva}, and is often augmented with depth or point-map observations~\cite{yu2024viewcrafter,mark2025trajectorycrafter,tu2025playerone} or joint RGB–geometry generation~\cite{wu2025spmem,zhen2025tesseract,zhu2025aether} to enforce 3D consistency.
More recent work expands the action space to include human body motion~\cite{bai2025peva,tu2025playerone} and humanoid control~\cite{mereu20251xworldmodel}. These approaches commonly adopt an image-to-video paradigm that hallucinates both the scene and its evolution from a single frame, entangling background synthesis with action-driven dynamics. Meanwhile, world models designed for robot-arm manipulation~\cite{guo2024pad,zhu2025uwm,jang2025dreamgen} typically assume a fixed camera viewpoint, and therefore cannot model tasks that require both navigation and manipulation.
Beyond these approaches, video diffusion models have also been explored as simulators for training embodied agents in driving~\cite{hu2023gaia,russell2025gaia2}, game environments~\cite{valevski2025gamengen,alonso2024diamond,oasis2024}, or text-generated worlds~\cite{bruce2024genie}. 
However, these approaches rely on abstract or limited action spaces and do not model fine-grained, contact-rich human–scene interactions.
In contrast, our approach assumes a known static environment and focuses on modeling interaction-induced visual dynamics from dexterous hand actions, while preserving the static scene.

\paragraph{Interactive 3D Scene Reconstruction.}
\label{sec:rel_digital_twin}
Recent advances in 3D reconstruction~\cite{wang2024dust3r,wang2025vggt} have made dense geometry and camera pose recovery from images increasingly accessible, lowering the barrier to creating static 3D scenes. 
However, most existing methods still produce static, monolithic scene representations that lack object-level disentanglement. 
To enable interaction and editing, recent work explores decompositional neural reconstruction~\cite{wu2022objectsdf,ni2024phyrecon,ni2025dprecon,xia2025holoscene} and articulated scene modeling, either by estimating articulation structures~\cite{liu2023paris,weng2024digitaltwinart,liu2025singapo,mandireal2code} or reconstructing full geometry and appearance using 3D Gaussian Splatting~\cite{xia2025drawer,liu2025artgs}. 
These approaches often require additional supervision such as joint-axis estimates~\cite{qian20233doi,sun2024opdmulti} or detailed segmentation~\cite{kirillov2023sam}, limiting scalability and generalization.
In contrast, our approach focuses on modeling everyday dexterous human–object interactions directly from static 3D scenes. 
By conditioning a video diffusion model on egocentric camera motion and fine-grained hand manipulations, we simulate temporally coherent interaction dynamics without explicit articulation modeling, bringing visual interactivity to static digital twins.

\section{Method}
\label{sec:method}
\subsection{Formulation of Dexterous World Models}
\label{subsection:overview}

\begin{figure*}[t]
\centering
\includegraphics[width=0.9\linewidth, trim={0cm 0cm 1cm 1cm}]{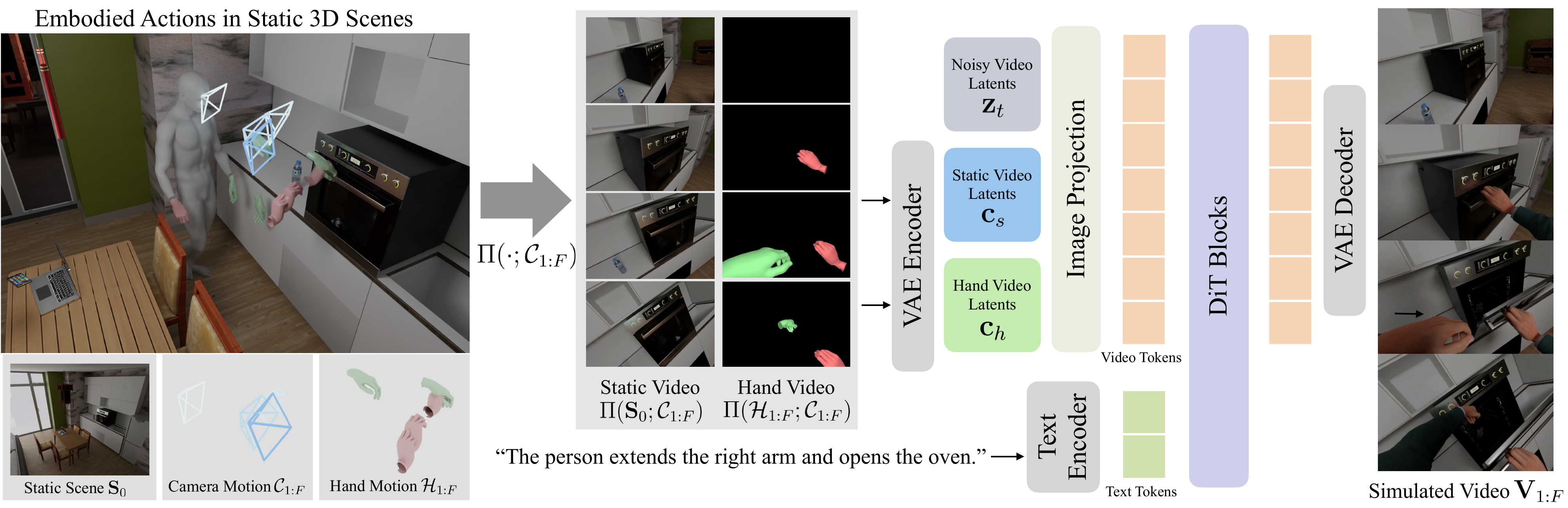}
\caption{\textbf{Overview.} DWM simulates egocentric visual dynamics induced by embodied actions within a given static 3D scene. We instantiate it as a video diffusion model conditioned on the egocentric projections of the static scene and hand trajectories.}
\label{fig:overview}
\end{figure*}

World models~\cite{ha2018worldmodel} aim to predict how actions transform the surrounding environment, allowing agents to simulate or anticipate future outcomes. A general formulation of a world model can be written as:
\begin{equation}
\label{eq:world_model}
\resizebox{0.91\linewidth}{!}{$
    p_{\theta}(\mathbf{S}_{1:F} \mid \mathbf{S}_0, \mathcal{A}_{1:F})
    =p_{\theta}\left( \left(\mathbf{S}_0 + \Delta \mathbf{S}_t \right)_{t={1:F}} \mid \mathbf{S}_0, \mathcal{A}_{1:F} \right),
$}
\end{equation}
where $\mathbf{S}_0$ denotes the initial scene environment and $\mathcal{A}_{1:F}$ represents the sequence actions applied to the world. $\mathbf{S}_{1:F}$ are the resulting environment states after the actions. To emphasize the dynamic component, we express each future state as $\mathbf{S}_t = \mathbf{S}_0 + \Delta \mathbf{S}_t$, where $\Delta \mathbf{S}_t$ captures the action-induced residual changes relative to the initial static scene.

Existing navigation-only visual world models~\cite{bar2025nwm,zhu2025aether} primarily focus on large-scale navigation and novel view prediction, where actions $\mathcal{A}_{1:F}$ mainly correspond to camera motion $\mathcal{C}_{1:F}$, that is
$p_{\theta}(\mathbf{V}_{1:F} \mid \mathbf{I}_{0}, \mathcal{C}_{1:F})$.
The $\mathbf{V}_{1:F}$ denotes future visual observations and $\mathbf{I}_{0}$ is the initial frame that provides a partial 2D observation of the underlying static scene $\mathbf{S}_0$. These models aim to predict unseen views $\mathbf{V}_{t}  = \Pi( \mathbf{S}_0 ; \mathcal{C}_{t})$, where $\Pi$ is a rendering function under camera view $\mathcal{C}_{t}$. This formulation often assumes a static environment with no state change, expressed as $\Delta \mathbf{S} = \varnothing$.

Alternatively, recent work~\cite{bai2025peva,tu2025playerone} attempts to incorporate human actions into dynamic scene modeling by expanding $\mathcal{A}_{1:F} = \{\mathcal{C}_{1:F}, \mathcal{H}_{1:F}\}$, where $\mathcal{H}$ denotes human actions. However, this formulation remains suboptimal for modeling dynamics because the model must synthesize both scene appearance and dynamic changes as an entangled representation. More formally, such models can be written as $ p_{\theta} \left(\mathbf{V}_{1:F} \mid \mathbf{I}_{0},  \mathcal{A}_{1:F}=\{ \mathcal{C}_{1:F}, \mathcal{H}_{1:F}\} \right)$, and it can be factorized by marginalizing over the latent static scene $\mathbf{S}_0$ and action-induced dynamics $\Delta \mathbf{S}_{1:F}$:
\begin{equation}
\label{eq:factor_world_model}
\resizebox{0.91\linewidth}{!}{$
\begin{aligned}
p_{\theta}(\mathbf{V}_{1:F} \mid \mathbf{I}_{0}, \mathcal{A}_{1:F})
= \int _{\mathbf{S}_0\, \Delta \mathbf{S}}
& p_{\theta}^{d}\big(
    \mathbf{S}_0, \Delta \mathbf{S}_{1:F} 
    \mid 
    \mathbf{I}_{0}, \mathcal{A}_{1:F}
\big) \\
& \cdot\,
p_{\theta}^{o}\big(
    \mathbf{V}_{1:F} 
    \mid 
    \mathbf{S}_0,  \Delta \mathbf{S}_{1:F}, \mathbf{I}_{0}, \mathcal{A}_{1:F}
\big)
\, d\mathbf{S}_0\, d\Delta \mathbf{S}_{1:F}.
\end{aligned}
$}
\end{equation}
Here, $p_{\theta}^{d}$ is the \emph{dynamics model} that samples the scene and action-induced changes $(\mathbf{S}_0, \Delta \mathbf{S}_{1:F})$ given the actions $\mathcal{A}_{1:F} = \{\mathcal{C}, \mathcal{H}\}_{1:F}$, while $p_{\theta}^{o}$ is the \emph{observation model} that maps latent world states to visual frames $\mathbf{V}_{1:F}$.
The observation model can be simplified to $p_{\theta}^{o}\big(
    \mathbf{V}_{1:F} \mid \mathbf{S}_0,  \Delta \mathbf{S}_{1:F}, \mathcal{C}_{1:F}
\big)$, as the visual outcome $\mathbf{V}_{1:F}$ is conditionally independent from the action input  $\{\mathcal{H}\}_{1:F}$ and image observation $\mathbf{I}_0$, given the known dynamics $\Delta \mathbf{S}_{1:F}$ and entire static scene $\mathbf{S}_0$. Similarly, in the dynamics model, we may drop the camera motion since it does not affect the world dynamics:  $p_{\theta}^{d}\big(
    \mathbf{S}_0, \Delta \mathbf{S}_{1:F} 
    \mid 
    \mathbf{I}_{0}, \mathcal{H}_{1:F}
\big).
$
However, since this dynamics model is still required to generate $\mathcal{S}_0$ as part of the output conditioned on the action sequence, it becomes vulnerable to altering the static world instead of focusing on the desired dynamic state changes $\Delta \mathbf{S}_{1:F}$.  
This coupling of scene generation and dynamics breaks causal consistency, making accurate dynamic modeling difficult.

In our DWM formulation, we explicitly separate the static world $\mathcal{S}_0$ from its dynamics by conditioning the model on $\mathcal{S}_0$.  
Our dexterous world model thus learns:
\begin{equation}
\label{eq:scene_world_model}
    p_{\theta}(\mathbf{V}_{1:F} \mid \mathbf{S}_0, \mathcal{A}_{1:F})
    = p_{\theta}(\mathbf{V}_{1:F} \mid \mathbf{S}_0, \{\mathcal{C}_{1:F}, \mathcal{H}_{1:F}\}),
\end{equation}
where $\mathbf{S}_0$ remains fixed and only its state evolves according to the applied actions.
By introducing temporal scene dynamics $\Delta \mathbf{S}_{1:F}$ as a latent variable and marginalizing over it, this can be expressed as
\begin{equation}
\label{eq:factor_scene_world_model}
\resizebox{0.91\linewidth}{!}{$
\begin{aligned}
p_{\theta}(\mathbf{V}_{1:F} \mid \mathbf{S}_0, \mathcal{A}_{1:F})
&= \int_{\Delta \mathbf{S}}
\, p_{\theta}^{d}\big(
    \Delta \mathbf{S}_{1:F} \mid \mathbf{S}_0, \mathcal{A}_{1:F}
\big)
\, p_{\theta}^{o}\big(
    \mathbf{V}_{1:F} \mid \mathbf{S}_0, \Delta \mathbf{S}_{1:F}, \mathcal{A}_{1:F}
\big)
\, d\Delta \mathbf{S} \\
&= \int_{\Delta \mathbf{S}}
\, p_{\theta}^{d}\big(
    \Delta \mathbf{S}_{1:F} \mid \mathbf{S}_0, \mathcal{H}_{1:F}
\big)
\, p_{\theta}^{o}\big(
    \mathbf{V}_{1:F} \mid \mathbf{S}_0, \Delta \mathbf{S}_{1:F}, \mathcal{C}_{1:F}
\big)
\, d\Delta \mathbf{S}.
\end{aligned}
$}
\end{equation}
Unlike the previous modeling in \cref{eq:factor_world_model}, our dynamics model $p_{\theta}^{d}$ only produces the dynamic changes $\Delta \mathbf{S}_{1:F}$ conditional to the action inputs $\mathcal{H}_{1:F}$.
The observation model $p_{\theta}^{o}$ then translates this evolving world, described by $(\mathbf{S}_0, \Delta \mathbf{S}_{1:F})$, into a sequence of visual frames observed along the camera trajectory $\mathcal{C}_{1:F}$, which is identical to the observation model in \cref{eq:factor_world_model}.
This structure defines a clear causal process: manipulation drives state transitions in the world, and the camera trajectory determines how these changes appear visually.

\subsection{Scene-Action-Conditioned Video Diffusion}
\label{subsection:method_diffusion}

Building on the formulation in \cref{eq:scene_world_model}, 
we instantiate our framework as an \emph{egocentric, scene-action-conditioned video diffusion model}.  
Given a static 3D scene $\mathbf{S}_0$ and a sequence of embodied actions $\mathcal{A}_{1:F}$, 
the model generates a temporally coherent video that visualizes plausible human-scene interactions with dynamic scene changes. We additionally condition the model on a text prompt $\mathcal{T}$, which supplies semantic guidance to the video diffusion process.
To ground this generative process in embodied perception,
we decompose $\mathcal{A}_{1:F}= \{\mathcal{C}_{1:F}, \mathcal{H}_{1:F}\}$ into a camera trajectory $\mathcal{C}_{1:F}$ and a hand manipulation trajectory $\mathcal{H}_{1:F}$, 
and assume a conditional independence structure such that the visual sequence $\mathcal{V}_{1:F}$ 
depends on the static world $\mathbf{S}_0$ and actions $\mathcal{A}_{1:F}$ only through their egocentric renderings:
\begin{equation}
\label{eq:ego_assumption}
\resizebox{0.91\linewidth}{!}{$
p(\mathbf{V}_{1:F} \mid \mathbf{S}_0, \mathcal{A}_{1:F}, \mathcal{T})
\approx
p(\mathbf{V}_{1:F} \mid \Pi(\mathbf{S}_0 ; \mathcal{C}_{1:F}), \Pi(\mathcal{H}_{1:F} ; \mathcal{C}_{1:F}), \mathcal{T}),
$}
\end{equation}
where $\Pi(\cdot ; \mathcal{C}_{1:F})$ denotes the 2D views observed along the egocentric camera trajectory $\mathcal{C}_{1:F}$.
This approximation grounds the dynamics in embodied perception, enforcing a causal relation where manipulation modifies the visual state of the scene, 
while the camera trajectory determines how those changes are observed.
Importantly, we instantiate the generative process as a latent video diffusion model~\cite{rombach2022ldm} conditioned on two egocentric signals:  
(1) a static-scene video, $\Pi(\mathbf{S}_0 ; \mathcal{C}_{1:F})$, that enforces spatial consistency along the specified camera trajectory, and  
(2) a egocentric hand-mesh rendering video~\cite{pavlakos2019smplx,romero2017mano}, $\Pi(\mathcal{H}_{1:F} ; \mathcal{C}_{1:F}))$, which encodes the manipulation actions that drive the dynamic scene changes. The model then simulates the resulting environment states induced by the hand action inputs under the same camera views,  $\mathbf{V}_{1:F} = \Pi( \mathbf{S}_0 + \Delta \mathbf{S}_{1:F} ; \mathcal{C}_{1:F}))$. See \cref{fig:overview} for the overview of our framework.

\paragraph{Inpainting Priors for Residual Dynamics Learning.}
We formulate the generation task as learning the residual visual change induced by hand manipulation action, rather than reconstructing the entire video from scratch.  
Given the target distribution 
$p(\mathbf{V}_{1:F} \mid \Pi(\mathbf{S}_0 ; \mathcal{C}_{1:F}), \Pi(\mathcal{H}_{1:F} ;  \mathcal{C}_{1:F}))$, 
we decompose it as:
\begin{equation}
\label{eq:residual}
\resizebox{0.91\linewidth}{!}{$
    \mathbf{V}_{1:F} = \Pi( \mathbf{S}_0 + \Delta \mathbf{S}_{1:F} ; \mathcal{C}_{1:F}))= \Pi(\mathbf{S}_0 ; \mathcal{C}_{1:F}) + \Delta \mathbf{V}_{1:F},
$}
\end{equation}
where $\Delta \mathbf{V}_{1:F}$ denotes the action-induced residual dynamics in 2D rendering space.  
Notably, when $\Delta \mathbf{S}_t$ affects only a small region or induces marginal changes, the corresponding $\Delta \mathbf{V}_t$ is similarly small, and the resulting frames remain close to the static rendering, $\Pi(\mathbf{S}_0 ; \mathcal{C}_{1:F})$.
Thus, an effective model for this task should combine an \emph{identity mapping} capability for preserving static regions with strong \emph{generative priors} for modeling dynamically changing areas.

Motivated by this observation, we reinterpret a pretrained video inpainting diffusion model~\cite{VideoX-Fun} as \emph{an identity function with generative prior} and adopt it as the initialization of our dexterous world model.  
Inpainting models are trained to reconstruct masked regions conditioned on visible context.
When the mask is set to $m=1$ (all pixels known), the model behaves as a near-identity operator that faithfully reproduces the input video.
Crucially, this is not a trivial identity: the pretrained inpainting network $\epsilon_{\theta_0}$ has learned to restore spatial structure, temporal smoothness, and appearance continuity, giving it a strong generative prior even under fully observed inputs. 
By initializing DWM from such an inpainting model, we encourage the diffusion process to preserve static scene appearance and camera motion, while learning to generate only the interaction-induced residual dynamics $\Delta \mathbf{V}_{1:F}$.
This formulation leverages the identity-preserving behavior of the inpainting model to stabilize training while learning interaction-induced residual dynamics.

\paragraph{Training.} The model operates in the latent space of a pretrained video variational autoencoder (VAE)~\cite{kingma2014vae,yang2025cogvideox}, 
which encodes an input video into latent tensors $\mathbf{z}_0 \in \mathbb{R}^{f \times c \times h \times w}$ 
following the encoder’s spatiotemporal compression ratio.  
At each diffusion timestep $t$, the Diffusion Transformer (DiT)~\cite{peebles2023dit} $\epsilon_{\theta}$
predicts the noise component from the noisy latent $\mathbf{z}_t$
conditioned on the static–scene and hand–mesh latents $(\mathbf{c}_s, \mathbf{c}_h)$.  
The training objective follows the standard latent diffusion loss~\cite{rombach2022ldm}:
\begin{equation}
\mathcal{L}_{\text{LDM}} 
= \mathbb{E}_{\mathbf{z}_0, t, \epsilon}
\!\left[
\left\|
\epsilon - \epsilon_\theta(\mathbf{z}_t, t \mid \mathbf{c}_s, \mathbf{c}_h)
\right\|_2^2
\right].
\end{equation}
The conditional latents are concatenated with $\mathbf{z}_t$ along the channel dimension before being processed by $\epsilon_\theta$.  
During inference, the model iteratively denoises $\mathbf{z}_t$ to obtain $\hat{\mathbf{z}}_0$, 
which is decoded by VAE into a realistic interaction video.

\subsection{Paired Interaction Video Dataset Construction}
\label{subsection:dataset}

Training our model requires paired data consisting of (i) a static scene video rendered along a camera trajectory, (ii) an egocentric hand-mesh video, and (iii) a corresponding interaction video depicting action-induced scene changes.
Collecting such triplets in real-world environments is challenging in practice, as it would require capturing both a static scene and a dynamic human–object interaction under a perfectly aligned egocentric camera trajectory.
This alignment is essential for learning how human actions transform a static scene while preserving spatial consistency.
Due to the practical difficulty of obtaining such paired data under dynamic egocentric viewpoints in the real world, we construct a hybrid training dataset.
Specifically, we combine synthetic egocentric interaction video pairs, which provide precise alignment, with fixed-camera real-world interaction videos, which offer realistic physical dynamics.
In addition, to evaluate generalization to real-world settings with dynamic viewpoints, we develop a dedicated data construction protocol to collect a real-world dataset for evaluation.

\paragraph{Synthetic Interaction-Static Scene Video Pairs.}
We leverage a synthetic 3D human-scene interaction dataset, TRUMANS~\cite{jiang2024trumans}, which offers full control over human actions, scene states, and camera trajectories, allowing exact alignment between static and interactive sequences.
The actor is parameterized using SMPL-X~\cite{pavlakos2019smplx}, which provides full-body pose control. We place a virtual camera at the midpoint between the eyes in the canonical pose, aligned with the head’s forward axis and rigidly attached to the head joint, to simulate egocentric viewpoints. For each pre-recorded action sequence in TRUMANS, we render three synchronized outputs: (1) the interaction video captures the full human–scene interaction from the moving egocentric camera; (2) the static-scene video is rendered by replaying the same camera trajectory over the static environment without applying any object state changes; and (3) the hand-mesh video is obtained by rendering only the hand surfaces, segmented from the full-body mesh, along the same trajectory. This setup ensures spatiotemporal alignment across all videos, providing a precise supervisory signal for learning manipulation-induced visual dynamics.

\paragraph{Fixed-Camera Real-World Interaction Videos.}
While synthetic data provides precise alignment, it lacks the complex physical effects observed in the real world, such as fluid dynamics, material deformation, and subtle visual variations. To incorporate such effects, we additionally use real-world human–object interaction videos recorded from fixed cameras~\cite{zhao2025taste}, where $\mathcal{C}_{t} = \mathcal{C}_0$ for all $t \in {1{:}F}$.
Under this setup, the egocentric rendering of the static scene is time-invariant:
\begin{equation}
    \Pi(\mathbf{S}_0 ; \mathcal{C}_t) = \Pi(\mathbf{S}_0 ; \mathcal{C}_0) = \mathbf{V}_0,
\quad \forall t \in \{1,\dots,F\}.
\end{equation}
Therefore, by simply repeating the first frame $\mathbf{V}_0$ across $F$ frames, we construct a static-scene video that aligns exactly with the interaction video in viewpoint. Combined with predicted hand-mesh renderings $\Pi(\mathcal{H}_{1:F} ; \mathcal{C}_{1:F})$, obtained via HaMeR~\cite{pavlakos2024hamer}, this enables us to construct training triplets $(\mathbf{V}_{1:F}, \Pi(\mathbf{S}_0 ; \mathcal{C}_{1:F}), \Pi(\mathcal{H}_{1:F} ; \mathcal{C}_{1:F}))$.
Notably, existing egocentric datasets~\cite{hoque2025egodex, damen2018scaling, zhao2025taste, fu2025gigahands} do not explicitly provide such pairs aligned under shared camera motion. Our construction leverages fixed-camera setups to approximate this pairing, enabling supervision without requiring separate static captures or 3D scene reconstruction.
This extends dexterous world modeling to real-world data while maintaining compatibility with our observation model in \cref{eq:ego_assumption}.

\paragraph{Real-World Captures for Evaluation.}
Although real-world paired data under dynamic egocentric viewpoints is impractical to obtain at scale, evaluating generalization to such settings is essential for validating a world model.
To this end, we develop a dedicated data construction protocol to collect a real-world dataset for evaluation under dynamic viewpoints.
We use Aria Glasses, which provide millimeter-level camera trajectory estimates via built-in SLAM~\cite{engel2023project}.
During capture, the operator first explores the scene before performing the interaction, enabling reconstruction of a 3D Gaussian representation of the static scene from pre-action frames~\cite{kerbl20233d,ye2025gsplat}.
The reconstructed scene is then rendered along the recorded camera trajectory during interaction, yielding paired static-scene videos aligned with the corresponding ground-truth interaction sequences.
Using this protocol, we collect 60 paired real-world samples under dynamic egocentric viewpoints, covering a wide range of interaction types, including simple and complex pick-and-place, articulated object manipulation (e.g., opening a washing machine, folding a chair), and counterfactual dynamics (e.g., pressing an elevator button and the door opens, turning on a faucet and water flows).
Examples are shown in our Supp. Mat..

\subsection{Action Evaluation with DWM}
\label{subsection:planning}
Given a static 3D scene $\mathbf{S}_0$ and a set of action candidates $\{\mathcal{A}^{(i)}\}_{i=1}^N$, we simulate their visual consequences using DWM as described in \cref{eq:scene_world_model}, producing videos $\{\mathbf{V}^{(i)}_{1:F}\}_{i=1}^N$. We then rank the actions according to how well their predicted visual outcomes align with a given goal, specified either as a language instruction or an image.
For language-based goals, we compute the semantic similarity between each generated video $\mathbf{V}^{(i)}_{1:F}$ and the goal text $g_\text{text}$ using a pretrained VideoCLIP model~\cite{wang2024videoclip}:
\begin{equation}
\label{eq:planning_text}
    s^{(i)}_{\text{text}} = \mathrm{sim}_\text{VC}(\mathbf{V}^{(i)}_{1:F}, g_\text{text}),
\end{equation}
where $\mathrm{sim}_\text{VC}(\cdot, \cdot)$ denotes the cosine similarity between video and text embeddings.
For image-based goals, we compare the final frame $\mathbf{I}^{(i)}_F = \mathbf{V}^{(i)}_F$ of each simulation against a goal image $\mathbf{I}_\text{goal}$ using the LPIPS perceptual metric~\cite{zhang2018unreasonable}:
\begin{equation}
\label{eq:planning_image}
    s^{(i)}_{\text{img}} = -\mathrm{LPIPS}(\mathbf{I}^{(i)}_F, \mathbf{I}_\text{goal}),
\end{equation}
where lower LPIPS indicates higher perceptual similarity.
The best action is selected as the one maximizing the score:
\begin{equation}
\label{eq:planning_argmax}
    \mathcal{A}^* = \arg\max_{\mathcal{A}^{(i)}} s^{(i)}_{\text{text}} \quad \text{or} \quad \mathcal{A}^* = \arg\max_{\mathcal{A}^{(i)}} s^{(i)}_{\text{img}},
\end{equation}
depending on the type of the given goal. This formulation enables goal-driven action selection via simulation, without requiring explicit reward functions or real-world trials.

\section{Experiments}
\label{sec:experiments}

\noindent \textbf{Dataset.}
We construct a benchmark of 144 samples, each consisting of a static scene video, a hand-mesh video, a ground-truth interaction video. To ensure comprehensive evaluations, the benchmark includes both synthetic and real-world data.
For the synthetic subset (Synthetic Dynamic), we sample 48 sequences from the TRUMANS~\cite{jiang2024trumans} that are excluded from training. The real-world subset includes 96 videos under static or dynamic camera settings. In the static camera setting (Real-World Static), we use 48 samples unseen during training from TASTE-Rob~\cite{zhao2025taste}. For the dynamic camera setting (Real-World Dynamic), we use 48 samples from our custom captured dataset by Aria Glasses.

\noindent \textbf{Baselines and Metrics.}
We compare our approach against two baselines that can directly utilize static scene videos as conditioning inputs.
First, We apply SDEdit~\cite{meng2021sdedit} to static videos using text-prompt guidance describing the target action and resulting scene changes (\textit{CVX SDEdit}). SDEdit first adds noise to the input video, then denoises it through the SDE prior under text conditioning for video editing. We use CogVideoX~\cite{yang2025cogvideox} for denoising and set the noise strength to 0.75 with 50 inference steps.
Second, we evaluate a fine-tune version of our base model, CogVideoX-Fun~\cite{VideoX-Fun} (\textit{CVX-Fun Fine-tuned}). CogVideoX-Fun is an inpainting-based video diffusion model that takes an input video, a binary mask sequence specifying the inpainting region, and a text prompt to generate inpainted videos. We set the mask as ones and fine-tune the model with our dataset without the hand-mesh video condition.
For the static camera setting, we additionally compare against InterDyn~\cite{akkerman2025interdyn}, which synthesizes hand–object interaction videos from an initial frame using a hand-mask video injected into the diffusion process via ControlNet~\cite{zhang2023adding}.

We evaluate generated videos using both perceptual and pixel-level metrics against ground-truth videos. Specifically, we report LPIPS~\cite{zhang2018unreasonable} and DreamSim~\cite{fu2023dreamsim} scores for perceptual similarity, and PSNR and SSIM for pixel-level quality. For each sample in our benchmark, we generate three videos with random seeds and report the averaged results.

\begin{table*}[t]
\centering
\resizebox{0.85\linewidth}{!}{  
\begin{tabular}{lcccccccccccc}
\toprule
& \multicolumn{4}{c}{\textbf{Synthetic}} & \multicolumn{8}{c}{\textbf{Real-World}} \\
\cmidrule(lr){2-5} \cmidrule(lr){6-13}
& \multicolumn{4}{c}{\textbf{Dynamic Camera}} & \multicolumn{4}{c}{\textbf{Static Camera}} & \multicolumn{4}{c}{\textbf{Dynamic Camera}} \\
\cmidrule(lr){2-5} \cmidrule(lr){6-9} \cmidrule(lr){10-13}
& PSNR$\uparrow$ & SSIM$\uparrow$ & LPIPS$\downarrow$ & DreamSim$\downarrow$
& PSNR$\uparrow$ & SSIM$\uparrow$ & LPIPS$\downarrow$ & DreamSim$\downarrow$
& PSNR$\uparrow$ & SSIM$\uparrow$ & LPIPS$\downarrow$ & DreamSim$\downarrow$ \\
\midrule
CVX SDEdit & 19.424 & 0.675 & 0.464 & 0.257 & 16.194 & 0.586 & 0.446 & 0.224 & 19.154 & 0.507 & 0.676 & 0.492 \\
CVX-Fun Fine-tuned & 20.541 & 0.767 & 0.370 & 0.175 & 18.951 & 0.780 & 0.265 & 0.089 & 18.129 & 0.472 & 0.591 & 0.328 \\
InterDyn & -- & -- & -- & -- & 19.331 & 0.744 & 0.240 & 0.135 & -- & -- & -- & -- \\
Ours & \textbf{25.031} & \textbf{0.844} & \textbf{0.289} & \textbf{0.086} & \textbf{21.547} & \textbf{0.816} & \textbf{0.227} & \textbf{0.057} & \textbf{21.654} & \textbf{0.550} & \textbf{0.557} & \textbf{0.225} \\
\bottomrule
\end{tabular}
}
\caption{\textbf{Quantitative comparisons on synthetic and real-world datasets.} 
Our method consistently achieves the best performance across all metrics and settings, demonstrating superior realism and physical coherence in scene-dynamics simulations.}

\label{tab:main}
\vspace{-3mm}
\end{table*}

\begin{figure*}[t]
\centering
\includegraphics[width=0.9\linewidth, trim={1.5cm 0cm 0cm 0cm}]{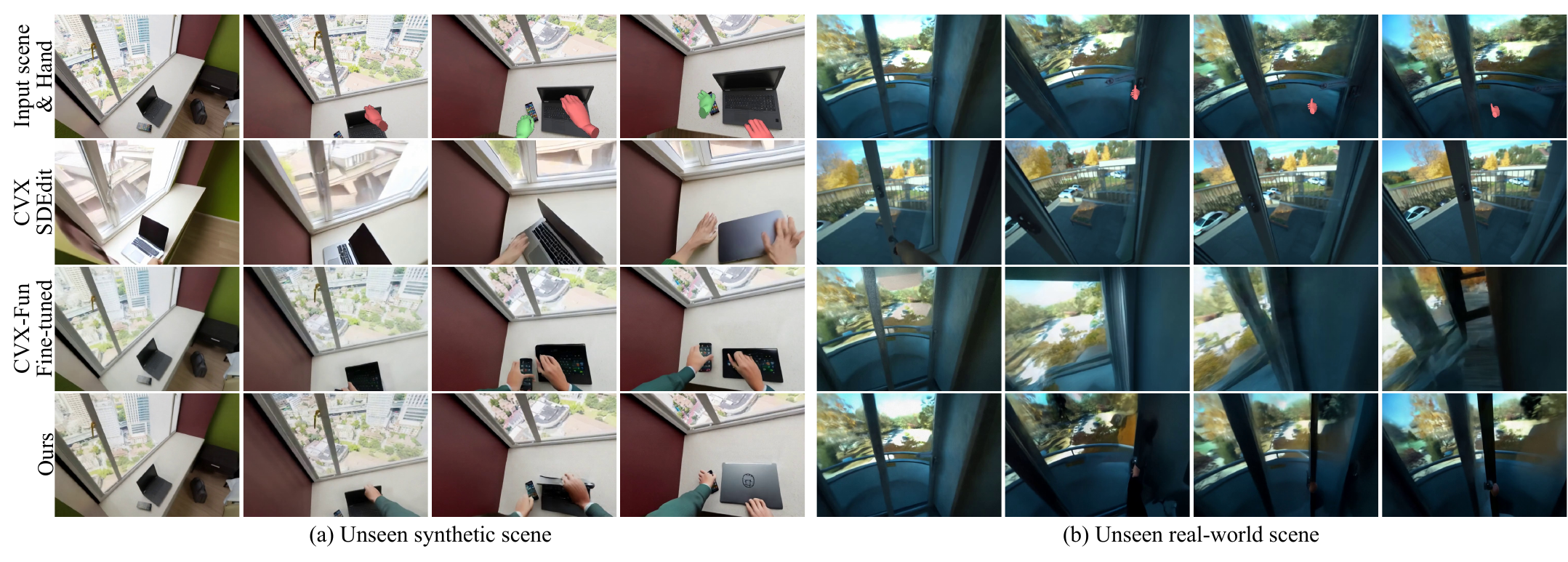}
\caption{\textbf{Qualitative comparison on synthetic and real-world scenes with dynamic view.} DWM successfully generates physically plausible simulations with dynamic view changes corresponding to the input hand actions. Notably, our method generalizes well to completely unseen real-world scenes, producing coherent action-conditioned dynamics such as opening a sliding window.}
\vspace{-3mm}
\label{fig:comp_dyna}
\end{figure*}

\begin{figure}
\centering
\includegraphics[width=0.9\columnwidth, trim={1.5cm 0cm 0cm 0cm}]{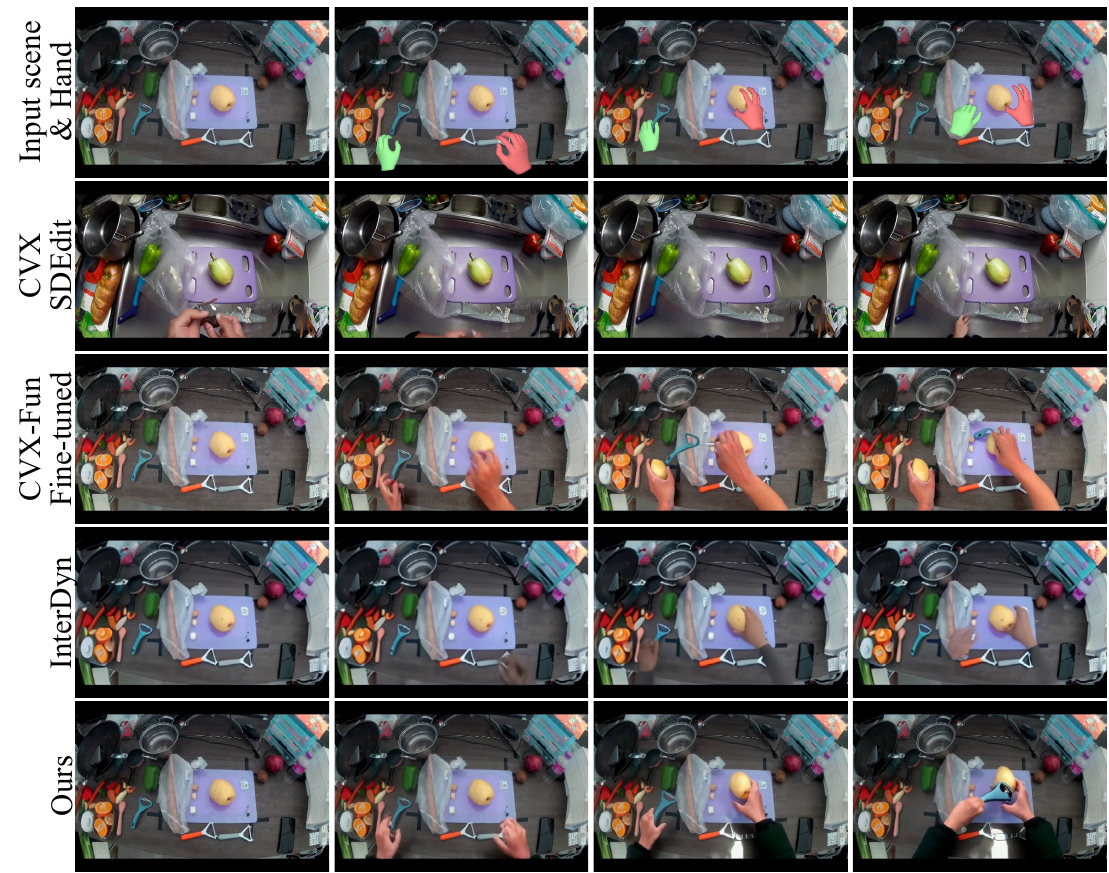}
\caption{\textbf{Qualitative comparison on real-world scenes with static camera.}
Our method produces realistic interactions with consistent scene dynamics. Baselines fail to perform meaningful actions or hallucinate incorrect interactions.}
\label{fig:comp_stat}
\vspace{-5mm}
\end{figure}

\begin{figure}[t]
\centering
\includegraphics[width=0.9\columnwidth, trim={1.25cm 0cm 0cm 0cm}]{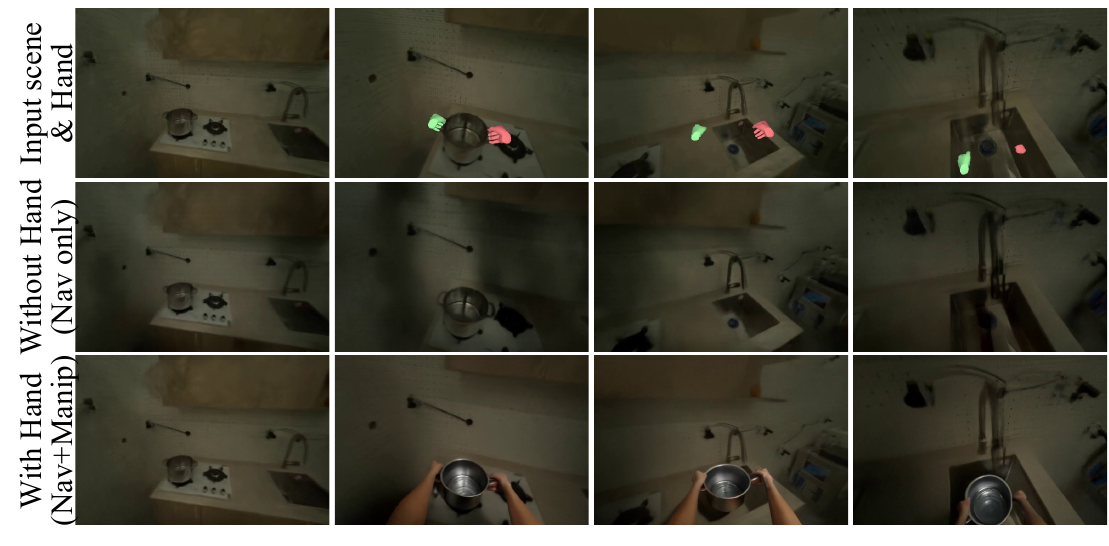}
\caption{\textbf{Navigation-manipulation disentanglement.} Without hand-motion input, DWM simulates navigation only.
Conditioning on hand motion enables the model to generate action-induced visual dynamics, highlighting navigation-manipulation disentanglement.}
\vspace{-5mm}
\label{fig:disentangle}
\end{figure}

\subsection{Comparison}
\label{sec:qual}
\cref{tab:main} shows quantitative performance across synthetic and real-world datasets under static or dynamic cameras, complementing the qualitative results presented in \cref{fig:comp_stat,fig:comp_dyna}.

\noindent \textbf{Real-World Scene with Static Camera.}
~\cref{fig:comp_stat} compares our method with baselines in a real-world scene with a static camera. CVX-SDEdit struggles to generate meaningful interactions while maintaining scene appearance. CVX-Fun Fine-tuned often fails to interact with the correct target and hallucinates the object of interest. InterDyn~\cite{akkerman2025interdyn} produces spatially aligned hands with the input masks but fails to model resulting object dynamics.
In contrast, our model synthesizes physically plausible interactions in which hand actions induce consistent scene changes, such as object displacement or articulation. Quantitatively, ours outperforms baselines across all metrics on real-world static camera videos.

\noindent \textbf{Scenes with Dynamic Camera.}
~\cref{fig:comp_dyna} shows comparisons on synthetic and real-world scenes captured in dynamic views. Despite real-world scenes with dynamic view being completely unseen during training, and the absence of any training samples involving opening a window, DWM successfully generates coherent simulations, demonstrating strong generalization. This strong generalization is quantitatively validated in \cref{tab:main}, where our method achieves superior performance across all metrics.

\subsection{Qualitative Results}

\noindent \textbf{Navigation-Manipulation Disentanglement.}
~\cref{fig:disentangle} demonstrates DWM’s ability to disentangle navigation and manipulation. Without hand motion conditioning, DWM operates as a pure navigator. When conditioned on hand motion, it additionally models manipulation-induced dynamics, generating realistic scene changes driven by hand actions.

\begin{table*}
\centering
\resizebox{0.9\linewidth}{!}{  
\begin{tabular}{lcccccccccccc}
\toprule
& \multicolumn{4}{c}{\textbf{Synthetic}} & \multicolumn{8}{c}{\textbf{Real-World}} \\
\cmidrule(lr){2-5} \cmidrule(lr){6-13}
& \multicolumn{4}{c}{\textbf{Dynamic Camera}} & \multicolumn{4}{c}{\textbf{Static Camera}} & \multicolumn{4}{c}{\textbf{Dynamic Camera}} \\
\cmidrule(lr){2-5} \cmidrule(lr){6-9} \cmidrule(lr){10-13}
& PSNR$\uparrow$ & SSIM$\uparrow$ & LPIPS$\downarrow$ & DreamSim$\downarrow$
& PSNR$\uparrow$ & SSIM$\uparrow$ & LPIPS$\downarrow$ & DreamSim$\downarrow$
& PSNR$\uparrow$ & SSIM$\uparrow$ & LPIPS$\downarrow$ & DreamSim$\downarrow$ \\
\midrule
TRUMANS Only & 24.151 & 0.834 & 0.304 & 0.093 & 17.959 & 0.766 & 0.304 & 0.124 & 20.654 & 0.520 & \textbf{0.543} & 0.273 \\
Ours & \textbf{25.031} & \textbf{0.844} & \textbf{0.289} & \textbf{0.086} & \textbf{21.547} & \textbf{0.816} & \textbf{0.227} & \textbf{0.057} & \textbf{21.654} & \textbf{0.550} & 0.557 & \textbf{0.225} \\
\bottomrule
\end{tabular}
}
\caption{\textbf{Ablation on training data composition.} 
Including real-world data for training significantly improves both perceptual and pixel-level metrics across synthetic and real-world test sets, demonstrating that static-camera real-world interactions enhance generalization.}
\vspace{-3mm}
\label{tab:ablation_dataset}
\end{table*}

\begin{table}
\centering
\resizebox{0.9\columnwidth}{!}{  
\begin{tabular}{lcccc}
\toprule
& PSNR$\uparrow$ & SSIM$\uparrow$ & LPIPS$\downarrow$ & DreamSim.$\downarrow$\\
\midrule
AdaLN (global) & 21.962 & 0.806 & 0.306 & 0.127 \\
AdaLN (per frame) & 22.789 & 0.827 & \textbf{0.288} & 0.110 \\
Hand Mask & 22.876 & 0.797 & 0.338 & 0.137 \\
Ours &  \textbf{24.151} & \textbf{0.834} & 0.304 & \textbf{0.094}  \\
\bottomrule
\end{tabular}
}
\caption{\textbf{Ablation on hand-motion conditioning.} 
We demonstrate the effectiveness of spatially aligned hand-mesh conditioning.}
\label{tab:ablation_conditioning}
\end{table}

\begin{figure}[t]
\centering
\includegraphics[width=0.9\columnwidth, trim={0cm 0cm 0cm 0cm}]{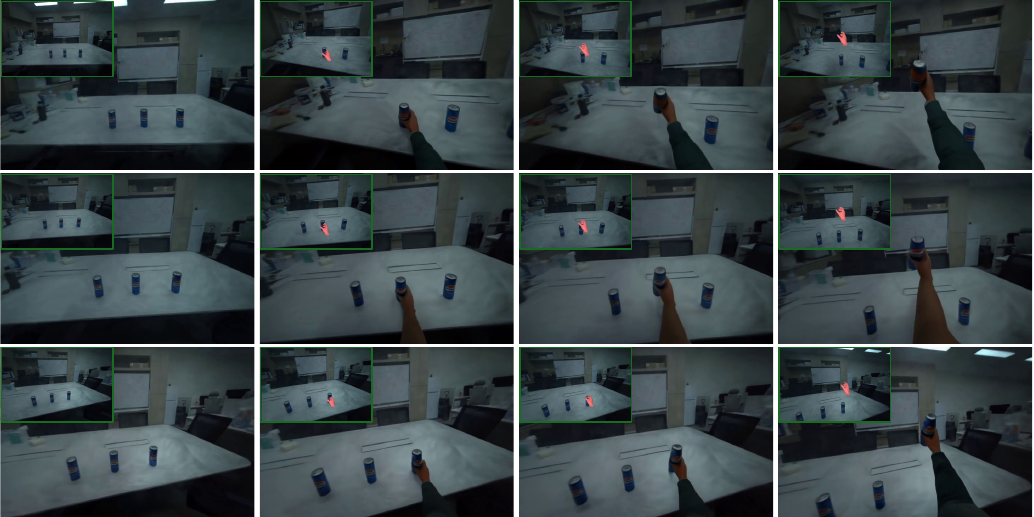}
\caption{\textbf{Accurate targeting.} DWM accurately manipulates the intended target object based on variations in hand-action condition. Input scene and hand conditions are shown in the top-left corners.}
\label{fig:complex}
\end{figure}
\noindent \textbf{Accurate Targeting.}
\cref{fig:complex} demonstrates DWM’s ability to precisely select and manipulate the target according to different hand-action conditions. The model responds to variations in hand motion, resulting in accurate interactions.

\begin{figure}[t]
\centering
\includegraphics[width=0.9\columnwidth, trim={0cm 0cm 0.8cm 0cm}]{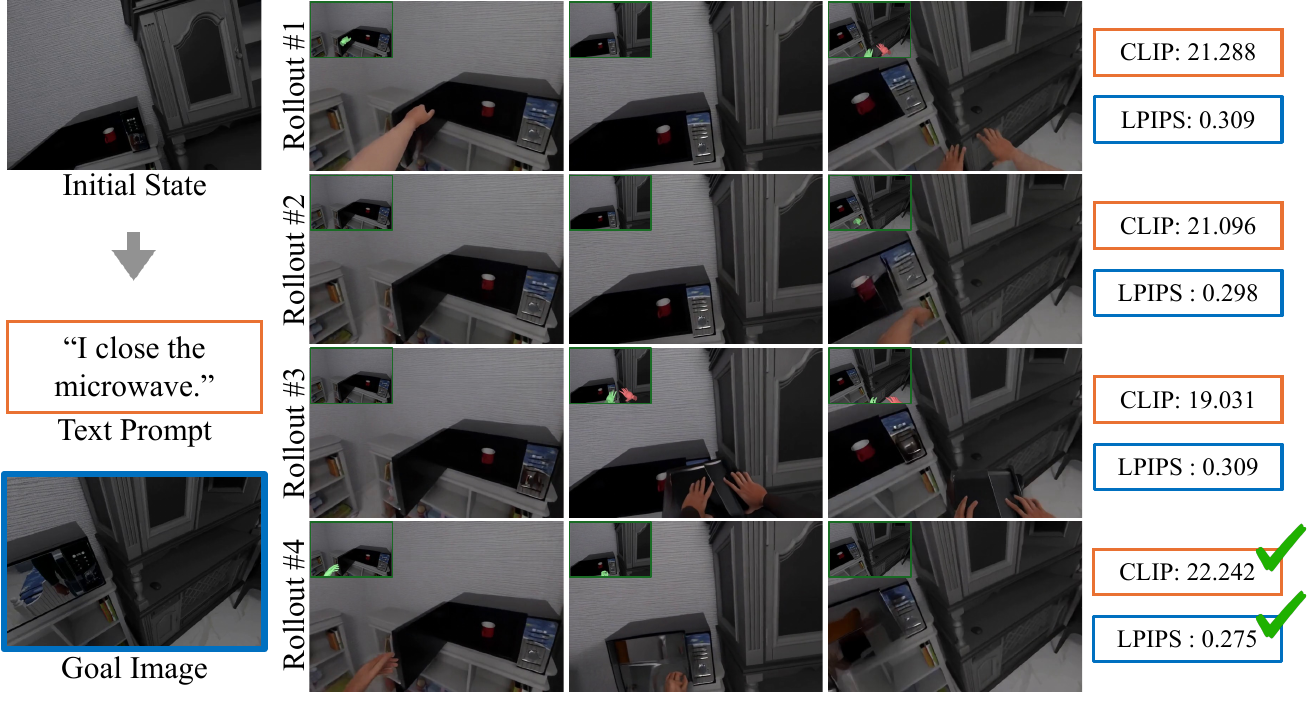}
\caption{\textbf{Action evaluation.} DWM simulates multiple candidate actions and ranks them based on goal alignment.}
\vspace{-5mm}
\label{fig:planning}
\end{figure}

\noindent \textbf{Action Evaluation.}
~\cref{fig:planning} illustrates how DWM can be used as an action evaluator. The model simulates outcomes for multiple action candidates and scores each based on its proximity to the desired goal. Depending on the goal specification, we measure either semantic alignment with VideoCLIP~\cite{wang2024videoclip} (for text goals) or perceptual similarity using LPIPS~\cite{zhang2018unreasonable} (for image goals). The action with the highest CLIP similarity or lowest LPIPS distance is selected.

\subsection{Ablation Study}

\noindent \textbf{Impact of Hybrid Dataset.}
We evaluate the impact of hybrid training data composition by comparing models trained with only TRUMANS versus those trained with both TRUMANS and TASTE-Rob. As shown in \cref{tab:ablation_dataset}, incorporating real-world data consistently improves performance on most metrics, not only on synthetic test sequences but also on Aria-captured real-world data, which includes both locomotion and fine-grained manipulation under dynamic camera motion. This is notable given that the real-world training data includes only static-camera interactions, demonstrating strong generalization to more complex real-world scenarios. The results validate our training data combination strategy: synthetic data offers controllable paired supervision with diverse camera motion, while real-world videos provide high-fidelity physical dynamics and visual realism.

\noindent \textbf{Hand Motion Conditioning.}
We compare three types of hand-motion conditioning: parameter-based modulation(AdaLN~\cite{peebles2023dit}), hand mask conditioning, and our hand-mesh-rendered conditioning (\cref{tab:ablation_conditioning}).
AdaLN-based conditioning directly injects MANO~\cite{romero2017mano} pose parameters with global orientation and translation of both hands into the diffusion transformer through Adaptive Layer Normalization.
We explore two variants: \textit{AdaLN (Global)}, which aggregates pose features over the entire sequence to provide a single global embedding, and \textit{AdaLN (Per-frame)}, which maintains frame-level embeddings to capture temporal variations.  
While both allow structured control, their parameter-space conditioning lacks pixel alignment, limiting the model’s ability to reproduce detailed hand–scene contact and appearance changes.  
\textit{Hand mask} instead provides pixel-level alignment by encoding binary hand segmentation videos via the same VAE encoder.  
However, the coarse silhouettes convey only location cues and lack precise articulation, leading to less coherent motion.  
\textit{Ours} uses rendered hand-mesh~\cite{pavlakos2019smplx,romero2017mano} videos as pixel-aligned conditioning signals.  
Unlike segmentation masks, the rendered meshes provide explicit articulation and contact geometry, allowing the model to directly associate hand motion with appearance changes in the scene.  
This explicit and spatially consistent conditioning leads to clear improvements, as shown in \cref{tab:ablation_conditioning}.

\section{Conclusion}
\label{sec:conclusion}
We introduced \textbf{Dexterous World Models (DWM)}, a video diffusion framework for modeling egocentric visual dynamics induced by locomotion and dexterous hand interactions in a known static 3D scene. Built on an egocentric observation model, DWM predicts residual visual changes conditioned on rendered static scene views and hand mesh trajectories. To train this model, we proposed a hybrid data construction strategy combining synthetic sequences with fixed-camera real-world videos, enabling aligned supervision and physically plausible dynamics. Experiments show that DWM generates realistic and grounded outcomes, generalizes to unseen real-world sequences involving both navigation and manipulation, and effectively disentangles viewpoint shifts from manipulation-induced state changes.

{
    \small
    \bibliographystyle{ieeenat_fullname}
    \bibliography{main}

@string{cvpr   = "Proc. CVPR"}

@string{iccv   = "Proc. ICCV"}

@string{eccv   = "Proc. ECCV"}

@string{threedv = "Proc. Intl. Conf. on 3D Vision (3DV)"}

@string{nips    = "NeurIPS"}

@string{neurips    = "NeurIPS"}

@string{iclr    = "Proc. ICLR"}

@string{icml    = "Proc. ICML"}

@string{tog     = "ACM Trans. Graph."}

@inproceedings{kingma2014vae,
  author= {Diederik P. Kingma and
                  Max Welling},
  title        = {Auto-Encoding Variational Bayes},
  booktitle    = ICLR,
  year         = {2014},
}

@inproceedings{pavlakos2024hamer,
  title={Reconstructing hands in 3d with transformers},
  author={Pavlakos, Georgios and Shan, Dandan and Radosavovic, Ilija and Kanazawa, Angjoo and Fouhey, David and Malik, Jitendra},
  booktitle=cvpr,
  year={2024}
}

@article{romero2017mano,
  title={Embodied hands: modeling and capturing hands and bodies together},
  author={Romero, Javier and Tzionas, Dimitrios and Black, Michael J},
  journal=tog,
  volume={36},
  number={6},
  pages={1--17},
  year={2017},
}

@article{ye2025gsplat,
  title={gsplat: An open-source library for Gaussian splatting},
  author={Ye, Vickie and Li, Ruilong and Kerr, Justin and Turkulainen, Matias and Yi, Brent and Pan, Zhuoyang and Seiskari, Otto and Ye, Jianbo and Hu, Jeffrey and Tancik, Matthew and others},
  journal={Journal of Machine Learning Research},
  volume={26},
  number={34},
  pages={1--17},
  year={2025}
}

@inproceedings{pavlakos2019smplx,
  title={Expressive body capture: 3d hands, face, and body from a single image},
  author={Pavlakos, Georgios and Choutas, Vasileios and Ghorbani, Nima and Bolkart, Timo and Osman, Ahmed AA and Tzionas, Dimitrios and Black, Michael J},
  booktitle=iccv,
  year={2019}
}

@inproceedings{jiang2024trumans,
  title={Scaling up dynamic human-scene interaction modeling},
  author={Jiang, Nan and Zhang, Zhiyuan and Li, Hongjie and Ma, Xiaoxuan and Wang, Zan and Chen, Yixin and Liu, Tengyu and Zhu, Yixin and Huang, Siyuan},
  booktitle=iccv,
  year={2024}
}

@inproceedings{zhao2025taste,
  title={TASTE-Rob: Advancing video generation of task-oriented hand-object interaction for generalizable robotic manipulation},
  author={Zhao, Hongxiang and Liu, Xingchen and Xu, Mutian and Hao, Yiming and Chen, Weikai and Han, Xiaoguang},
  booktitle=cvpr,
  year={2025}
}

@inproceedings{rombach2022ldm,
  title={High-resolution image synthesis with latent diffusion models},
  author={Rombach, Robin and Blattmann, Andreas and Lorenz, Dominik and Esser, Patrick and Ommer, Bj{\"o}rn},
  booktitle=cvpr,
  year={2022}
}

@inproceedings{
yang2025cogvideox,
title={CogVideoX: Text-to-Video Diffusion Models with An Expert Transformer},
author={Zhuoyi Yang and Jiayan Teng and Wendi Zheng and Ming Ding and Shiyu Huang and Jiazheng Xu and Yuanming Yang and Wenyi Hong and Xiaohan Zhang and Guanyu Feng and Da Yin and Yuxuan.Zhang and Weihan Wang and Yean Cheng and Bin Xu and Xiaotao Gu and Yuxiao Dong and Jie Tang},
booktitle=iclr,
year={2025},
}

@inproceedings{peebles2023dit,
  title={Scalable diffusion models with transformers},
  author={Peebles, William and Xie, Saining},
  booktitle=iccv,
  year={2023}
}

@inproceedings{hafner2019planet,
  title={Learning latent dynamics for planning from pixels},
  author={Hafner, Danijar and Lillicrap, Timothy and Fischer, Ian and Villegas, Ruben and Ha, David and Lee, Honglak and Davidson, James},
  booktitle=icml,
  year={2019},
}

@article{hafner2025dreamerv3,
  title={Mastering diverse control tasks through world models},
  author={Hafner, Danijar and Pasukonis, Jurgis and Ba, Jimmy and Lillicrap, Timothy},
  journal={Nature},
  pages={1--7},
  year={2025},
}

@misc{zhou2024dinowmworldmodelspretrained,
      title={DINO-WM: World Models on Pre-trained Visual Features enable Zero-shot Planning}, 
      author={Gaoyue Zhou and Hengkai Pan and Yann LeCun and Lerrel Pinto},
      year={2024},
      eprint={2411.04983},
      archivePrefix={arXiv},
      primaryClass={cs.RO},
      url={https://arxiv.org/abs/2411.04983}, 
}

@inproceedings{bar2025nwm,
  title={Navigation world models},
  author={Bar, Amir and Zhou, Gaoyue and Tran, Danny and Darrell, Trevor and LeCun, Yann},
  booktitle={Proceedings of the Computer Vision and Pattern Recognition Conference},
  pages={15791--15801},
  year={2025}
}

@article{mereu20251xworldmodel,
  title={Generative World Modelling for Humanoids: 1X World Model Challenge Technical Report},
  author={Mereu, Riccardo and Scannell, Aidan and Hou, Yuxin and Zhao, Yi and Jitta, Aditya and Dominguez, Antonio and Acerbi, Luigi and Storkey, Amos and Chang, Paul},
  journal={arXiv preprint arXiv:2510.07092},
  year={2025}
}

@article{hu2023gaia,
  title={Gaia-1: A generative world model for autonomous driving},
  author={Hu, Anthony and Russell, Lloyd and Yeo, Hudson and Murez, Zak and Fedoseev, George and Kendall, Alex and Shotton, Jamie and Corrado, Gianluca},
  journal={arXiv preprint arXiv:2309.17080},
  year={2023}
}

@article{russell2025gaia2,
  title={Gaia-2: A controllable multi-view generative world model for autonomous driving},
  author={Russell, Lloyd and Hu, Anthony and Bertoni, Lorenzo and Fedoseev, George and Shotton, Jamie and Arani, Elahe and Corrado, Gianluca},
  journal={arXiv preprint arXiv:2503.20523},
  year={2025}
}

@inproceedings{zhu2025aether,
  title={Aether: Geometric-Aware Unified World Modeling},
  author={Zhu, Haoyi and Wang, Yifan and Zhou, Jianjun and Chang, Wenzheng and Zhou, Yang and Li, Zizun and Chen, Junyi and Shen, Chunhua and Pang, Jiangmiao and He, Tong},
  booktitle=iccv,
  year={2025}
}

@inproceedings{bruce2024genie,
  title={Genie: Generative interactive environments},
  author={Bruce, Jake and Dennis, Michael D and Edwards, Ashley and Parker-Holder, Jack and Shi, Yuge and Hughes, Edward and Lai, Matthew and Mavalankar, Aditi and Steigerwald, Richie and Apps, Chris and others},
  booktitle=icml,
  year={2024}
}

@inproceedings{
valevski2025gamengen,
title={Diffusion Models Are Real-Time Game Engines},
author={Dani Valevski and Yaniv Leviathan and Moab Arar and Shlomi Fruchter},
booktitle=iclr,
year={2025},
}

@article{oasis2024,
  author    = {Decart and Julian Quevedo and Quinn McIntyre and Spruce Campbell and Xinlei Chen and Robert Wachen},
  title     = {Oasis: A Universe in a Transformer},
  year      = {2024},
  url       = {https://oasis-model.github.io/}
}

@article{alonso2024diamond,
  title={Diffusion for world modeling: Visual details matter in atari},
  author={Alonso, Eloi and Jelley, Adam and Micheli, Vincent and Kanervisto, Anssi and Storkey, Amos J and Pearce, Tim and Fleuret, Fran{\c{c}}ois},
  journal=neurips,
  year={2024}
}

@article{zhou2025seva,
  title={Stable virtual camera: Generative view synthesis with diffusion models},
  author={Zhou, Jensen and Gao, Hang and Voleti, Vikram and Vasishta, Aaryaman and Yao, Chun-Han and Boss, Mark and Torr, Philip and Rupprecht, Christian and Jampani, Varun},
  journal={arXiv preprint arXiv:2503.14489},
  year={2025}
}

@article{yu2024viewcrafter,
  title={Viewcrafter: Taming video diffusion models for high-fidelity novel view synthesis},
  author={Yu, Wangbo and Xing, Jinbo and Yuan, Li and Hu, Wenbo and Li, Xiaoyu and Huang, Zhipeng and Gao, Xiangjun and Wong, Tien-Tsin and Shan, Ying and Tian, Yonghong},
  journal={arXiv preprint arXiv:2409.02048},
  year={2024}
}

@inproceedings{mark2025trajectorycrafter,
  title={Trajectorycrafter: Redirecting camera trajectory for monocular videos via diffusion models},
  author={YU, Mark  and Hu, Wenbo and Xing, Jinbo and Shan, Ying},
  booktitle=iccv,
  year={2025}
}

@article{zhen2025tesseract,
  title={TesserAct: learning 4D embodied world models},
  author={Zhen, Haoyu and Sun, Qiao and Zhang, Hongxin and Li, Junyan and Zhou, Siyuan and Du, Yilun and Gan, Chuang},
  journal={arXiv preprint arXiv:2504.20995},
  year={2025}
}

@article{ha2018worldmodel,
  title={Recurrent world models facilitate policy evolution},
  author={Ha, David and Schmidhuber, J{\"u}rgen},
  journal=nips,
  year={2018}
}

@article{bai2025peva,
  title={Whole-Body Conditioned Egocentric Video Prediction},
  author={Bai, Yutong and Tran, Danny and Bar, Amir and LeCun, Yann and Darrell, Trevor and Malik, Jitendra},
  journal={arXiv preprint arXiv:2506.21552},
  year={2025}
}

@article{tu2025playerone,
  title={PlayerOne: Egocentric World Simulator},
  author={Tu, Yuanpeng and Luo, Hao and Chen, Xi and Bai, Xiang and Wang, Fan and Zhao, Hengshuang},
  journal={arXiv preprint arXiv:2506.09995},
  year={2025}
}

@article{wu2025spmem,
  title={Video World Models with Long-term Spatial Memory},
  author={Wu, Tong and Yang, Shuai and Po, Ryan and Xu, Yinghao and Liu, Ziwei and Lin, Dahua and Wetzstein, Gordon},
  journal={arXiv preprint arXiv:2506.05284},
  year={2025}
}

@article{guo2024pad,
  title={Prediction with action: Visual policy learning via joint denoising process},
  author={Guo, Yanjiang and Hu, Yucheng and Zhang, Jianke and Wang, Yen-Jen and Chen, Xiaoyu and Lu, Chaochao and Chen, Jianyu},
  journal=neurips,
  year={2024}
}

@inproceedings{zhu2025uwm,
    author    = {Zhu, Chuning and Yu, Raymond and Feng, Siyuan and Burchfiel, Benjamin and Shah, Paarth and Gupta, Abhishek},
    title     = {Unified World Models: Coupling Video and Action Diffusion for Pretraining on Large Robotic Datasets},
    booktitle = {Proceedings of Robotics: Science and Systems (RSS)},
    year      = {2025},
}

@article{jang2025dreamgen,
  title={DreamGen: Unlocking Generalization in Robot Learning through Video World Models},
  author={Jang, Joel and Ye, Seonghyeon and Lin, Zongyu and Xiang, Jiannan and Bjorck, Johan and Fang, Yu and Hu, Fengyuan and Huang, Spencer and Kundalia, Kaushil and Lin, Yen-Chen and others},
  journal={arXiv preprint arXiv:2505.12705v2},
  year={2025}
}

@inproceedings{qian20233doi,
    title={Understanding 3D Object Interaction from a Single Image},
    author={Qian, Shengyi and Fouhey, David F},
    booktitle = iccv,
    year={2023}
}

@inproceedings{ni2025dprecon,
  title={Decompositional neural scene reconstruction with generative diffusion prior},
  author={Ni, Junfeng and Liu, Yu and Lu, Ruijie and Zhou, Zirui and Zhu, Song-Chun and Chen, Yixin and Huang, Siyuan},
  booktitle={Proceedings of the Computer Vision and Pattern Recognition Conference},
  pages={6022--6033},
  year={2025}
}

@inproceedings{xia2025drawer,
  title={Drawer: Digital reconstruction and articulation with environment realism},
  author={Xia, Hongchi and Su, Entong and Memmel, Marius and Jain, Arhan and Yu, Raymond and Mbiziwo-Tiapo, Numfor and Farhadi, Ali and Gupta, Abhishek and Wang, Shenlong and Ma, Wei-Chiu},
  booktitle=cvpr,
  year={2025}
}

@article{liu2025artgs,
  title={Artgs: Building interactable replicas of complex articulated objects via gaussian splatting},
  author={Liu, Yu and Jia, Baoxiong and Lu, Ruijie and Ni, Junfeng and Zhu, Song-Chun and Huang, Siyuan},
  journal={arXiv preprint arXiv:2502.19459},
  year={2025}
}

@inproceedings{liu2023paris,
  title={Paris: Part-level reconstruction and motion analysis for articulated objects},
  author={Liu, Jiayi and Mahdavi-Amiri, Ali and Savva, Manolis},
  booktitle={Proceedings of the IEEE/CVF International Conference on Computer Vision},
  pages={352--363},
  year={2023}
}

@inproceedings{weng2024digitaltwinart,
    title={Neural Implicit Representation for Building Digital Twins of Unknown Articulated Objects}, 
    author={Yijia Weng and Bowen Wen and Jonathan Tremblay and Valts Blukis and Dieter Fox and Leonidas Guibas and Stan Birchfield},
    booktitle=cvpr,
    year={2024}
}

@inproceedings{sun2024opdmulti,
  title={Opdmulti: Openable part detection for multiple objects},
  author={Sun, Xiaohao and Jiang, Hanxiao and Savva, Manolis and Chang, Angel},
  booktitle=threedv,
  year={2024},
}

@inproceedings{liu2025singapo,
  title={SINGAPO: Single Image Controlled Generation of Articulated Parts in Objects},
  author={Liu, Jiayi and Iliash, Denys and Chang, Angel X and Savva, Manolis and Amiri, Ali Mahdavi},
  booktitle=iclr,
  year={2025}
}

@inproceedings{mandireal2code,
  title={Real2Code: Reconstruct Articulated Objects via Code Generation},
  author={Mandi, Zhao and Weng, Yijia and Bauer, Dominik and Song, Shuran},
  booktitle=iclr,
  year={2025}
}

@article{xia2025holoscene,
  title={HoloScene: Simulation-Ready Interactive 3D Worlds from a Single Video},
  author={Xia, Hongchi and Lin, Chih-Hao and Hsu, Hao-Yu and Leboutet, Quentin and Gao, Katelyn and Paulitsch, Michael and Ummenhofer, Benjamin and Wang, Shenlong},
  journal={arXiv preprint arXiv:2510.05560},
  year={2025}
}

@inproceedings{wu2022objectsdf,
  title={Object-compositional neural implicit surfaces},
  author={Wu, Qianyi and Liu, Xian and Chen, Yuedong and Li, Kejie and Zheng, Chuanxia and Cai, Jianfei and Zheng, Jianmin},
  booktitle=eccv,
  year={2022},
}

@article{ni2024phyrecon,
  title={Phyrecon: Physically plausible neural scene reconstruction},
  author={Ni, Junfeng and Chen, Yixin and Jing, Bohan and Jiang, Nan and Wang, Bin and Dai, Bo and Li, Puhao and Zhu, Yixin and Zhu, Song-Chun and Huang, Siyuan},
  journal=neurips,
  year={2024}
}

@article{hoque2025egodex,
  title={EgoDex: Learning Dexterous Manipulation from Large-Scale Egocentric Video},
  author={Hoque, Ryan and Huang, Peide and Yoon, David J and Sivapurapu, Mouli and Zhang, Jian},
  journal={arXiv preprint arXiv:2505.11709},
  year={2025}
}

@inproceedings{damen2018scaling,
  title={Scaling egocentric vision: The epic-kitchens dataset},
  author={Damen, Dima and Doughty, Hazel and Farinella, Giovanni Maria and Fidler, Sanja and Furnari, Antonino and Kazakos, Evangelos and Moltisanti, Davide and Munro, Jonathan and Perrett, Toby and Price, Will and others},
  booktitle=eccv,
  year={2018}
}

@article{engel2023project,
  title={Project aria: A new tool for egocentric multi-modal ai research},
  author={Engel, Jakob and Somasundaram, Kiran and Goesele, Michael and Sun, Albert and Gamino, Alexander and Turner, Andrew and Talattof, Arjang and Yuan, Arnie and Souti, Bilal and Meredith, Brighid and others},
  journal={arXiv preprint arXiv:2308.13561},
  year={2023}
}

@article{kerbl20233d,
  title={3D Gaussian splatting for real-time radiance field rendering.},
  author={Kerbl, Bernhard and Kopanas, Georgios and Leimk{\"u}hler, Thomas and Drettakis, George},
  journal=tog,
  year={2023}
}

@misc{VideoX-Fun,
  Author = {aigc-apps},
  Note = {\url{https://https://github.com/aigc-apps/VideoX-Fun}},
  Year = {2024}
}

@article{hu2022lora,
  title={Lora: Low-rank adaptation of large language models.},
  author={Hu, Edward J and Shen, Yelong and Wallis, Phillip and Allen-Zhu, Zeyuan and Li, Yuanzhi and Wang, Shean and Wang, Lu and Chen, Weizhu and others},
  journal=iclr,
  year={2022}
}

@article{loshchilov2017decoupled,
  title={Decoupled weight decay regularization},
  author={Loshchilov, Ilya and Hutter, Frank},
  journal=iclr,
  year={2019}
}

@inproceedings{akkerman2025interdyn,
  title={InterDyn: Controllable interactive dynamics with video diffusion models},
  author={Akkerman, Rick and Feng, Haiwen and Black, Michael J and Tzionas, Dimitrios and Abrevaya, Victoria Fern{\'a}ndez},
  booktitle=cvpr,
  year={2025}
}

@inproceedings{zhang2023adding,
  title={Adding conditional control to text-to-image diffusion models},
  author={Zhang, Lvmin and Rao, Anyi and Agrawala, Maneesh},
  booktitle=iccv,
  year={2023}
}

@inproceedings{zhang2018unreasonable,
  title={The unreasonable effectiveness of deep features as a perceptual metric},
  author={Zhang, Richard and Isola, Phillip and Efros, Alexei A and Shechtman, Eli and Wang, Oliver},
  booktitle=cvpr,
  year={2018}
}

@article{fu2023dreamsim,
  title={Dreamsim: Learning new dimensions of human visual similarity using synthetic data},
  author={Fu, Stephanie and Tamir, Netanel and Sundaram, Shobhita and Chai, Lucy and Zhang, Richard and Dekel, Tali and Isola, Phillip},
  journal=neurips,
  year={2023}
}

@inproceedings{fu2025gigahands,
  title={Gigahands: A massive annotated dataset of bimanual hand activities},
  author={Fu, Rao and Zhang, Dingxi and Jiang, Alex and Fu, Wanjia and Funk, Austin and Ritchie, Daniel and Sridhar, Srinath},
  booktitle=cvpr,
  year={2025}
}

@inproceedings{wang2024dust3r,
  title={Dust3r: Geometric 3d vision made easy},
  author={Wang, Shuzhe and Leroy, Vincent and Cabon, Yohann and Chidlovskii, Boris and Revaud, Jerome},
  booktitle={Proceedings of the IEEE/CVF Conference on Computer Vision and Pattern Recognition},
  pages={20697--20709},
  year={2024}
}

@inproceedings{kirillov2023sam,
  title={Segment anything},
  author={Kirillov, Alexander and Mintun, Eric and Ravi, Nikhila and Mao, Hanzi and Rolland, Chloe and Gustafson, Laura and Xiao, Tete and Whitehead, Spencer and Berg, Alexander C and Lo, Wan-Yen and others},
  booktitle=iccv,
  year={2023}
}

@inproceedings{wang2025vggt,
  title={Vggt: Visual geometry grounded transformer},
  author={Wang, Jianyuan and Chen, Minghao and Karaev, Nikita and Vedaldi, Andrea and Rupprecht, Christian and Novotny, David},
  booktitle=cvpr,
  year={2025}
}

@article{wang2024videoclip,
  title={Videoclip-xl: Advancing long description understanding for video clip models},
  author={Wang, Jiapeng and Wang, Chengyu and Huang, Kunzhe and Huang, Jun and Jin, Lianwen},
  journal={arXiv preprint arXiv:2410.00741},
  year={2024}
}

@article{meng2021sdedit,
  title={Sdedit: Guided image synthesis and editing with stochastic differential equations},
  author={Meng, Chenlin and He, Yutong and Song, Yang and Song, Jiaming and Wu, Jiajun and Zhu, Jun-Yan and Ermon, Stefano},
  journal=iclr,
  year={2022}
}

@article{lepert2025masquerade,
    title={Masquerade: Learning from In-the-wild Human Videos using Data-Editing},
    author={Lepert, Marion and Fang, Jiaying and Bohg, Jeannette},
    journal={arXiv preprint arXiv:2508.09976},
    year={2025}
  }

@article{lepert2025phantomtrainingrobotsrobots,
        title={Phantom: Training Robots Without Robots Using Only Human Videos}, 
        author={Marion Lepert and Jiaying Fang and Jeannette Bohg},
        year={2025},
        eprint={2503.00779},
        archivePrefix={arXiv},
        primaryClass={cs.RO},
        url={https://arxiv.org/abs/2503.00779}, 
  }
}

\ifarxiv

\maketitlesupplementary

\appendix

\section{Implementation Details.}
We adopt CogVideoX-Fun-V1.5-5B-InP~\cite{VideoX-Fun} as our base video diffusion model, generating output simulations at a resolution of 720 × 480 with 49 frames. For model fine-tuning, we apply LoRA~\cite{hu2022lora} to the diffusion transformer with a rank of 64 and $\alpha=64$. During training, only the added LoRA layers and the image projection layer are optimized, while all other parameters remain frozen. We train the model using the AdamW~\cite{loshchilov2017decoupled} optimizer with a learning rate of $1\times10^{-4}$ and an effective batch size of 56. Training approximately takes 10 days on 4 NVIDIA A100 GPUs.

\section{Additional Qualitative Results}
We provide additional DWM-generated simulations in \cref{fig:extra_4}–\cref{fig:extra_3}, along with their corresponding input static scenes hand actions. As shown in \cref{fig:extra_4,fig:extra_5}, DWM demonstrates strong generalization capability, producing realistic and physically consistent interactions in unseen real-world environments with dynamic viewpoint changes, despite the absence of such data during training. \cref{fig:extra_1} presents additional results on the synthetic dataset under a static camera, while \cref{fig:extra_2,fig:extra_3} show additional results on the real-world dataset under a static camera.

\section{Real-World Dataset with Dynamic View}
To evaluate DWM under realistic embodied-view conditions, we collect a real-world dataset with dynamic egocentric camera motion. Since existing egocentric datasets do not provide paired static scene videos for dynamic views, we follow the capture protocol introduced in the main paper, using Aria Glasses to record accurate camera trajectories and reconstruct a static 3D Gaussian scene from pre-action frames. Rendering this reconstruction along the interaction trajectory produces the paired static–dynamic video pairs required by our task.
Our current dataset contains 48 paired samples covering diverse interactions, including pick-and-place, articulated object manipulation, and counterfactual dynamics. Examples are shown in \cref{fig:data_0,fig:data_1}.

\section{Ablation Study on Base Models}

\begin{figure}
\centering
    \includegraphics[width=1.0\columnwidth, trim={0.5cm 0cm 0cm 0cm}]{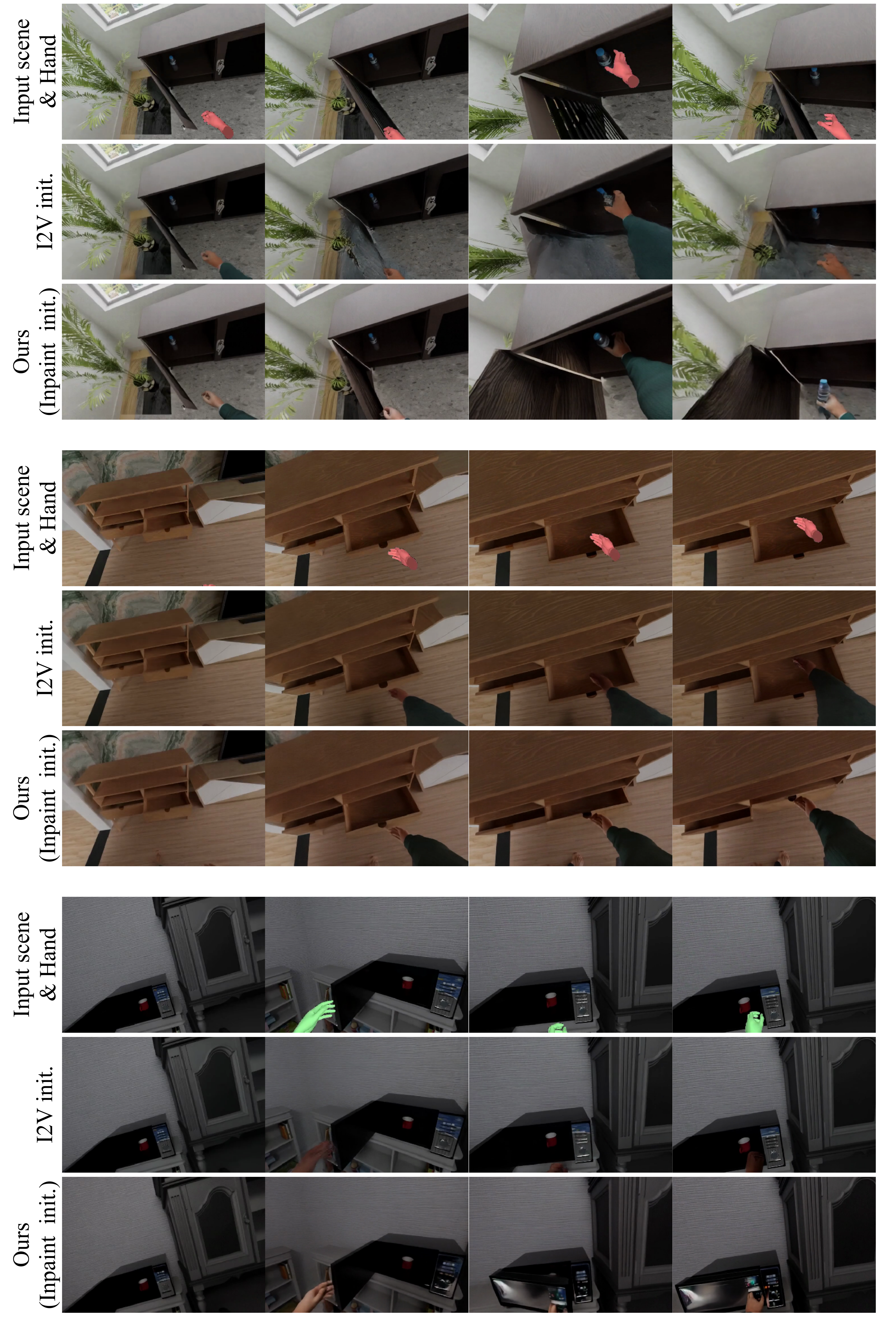}
    \caption{\textbf{Qualitative comparison for the base model ablations.} Our inpainting-based initialization consistently achieves better DreamSim scores compared to I2V initialization.}
\label{fig:ablation_base_model_qual}
\end{figure}
As discussed in \cref{subsection:method_diffusion}, we adopt the inpainting variant of CogVideo-X~\cite{VideoX-Fun} as our base model. 
This choice is motivated by the observation that a pretrained inpainting video diffusion model with a full mask $m=1$ (all pixels known) behaves as a near-identity operator while retaining a rich generative prior. 
To validate this hypothesis, we compare our model initialized from the inpainting model with the same architecture initialized from the Image-to-Video (I2V) model~\cite{yang2025cogvideox}. 
\Cref{tab:ablation_base_model} reports the DreamSim~\cite{fu2023dreamsim} metric evaluated every 1000 training steps up to 4000 iterations for the two models. 
The results show that the inpainting-based initialization achieves consistently better DreamSim scores across training iterations.
\Cref{fig:ablation_base_model_qual} further presents qualitative comparisons. 
While the I2V-initialized model struggles to properly capture action-conditioned dynamics, our inpainting-initialized model more effectively learns the motion patterns induced by manipulation. These observations provide further empirical support for our hypothesis that full-mask inpainting priors offer a more suitable initialization for residual dynamics learning than I2V-based initialization.

\begin{table}[t]
\centering
\small
\begin{tabular}{lccccc}
\toprule
{Initialize} & \multicolumn{5}{c}{{DreamSim$\downarrow$ @ Iterations}} \\
\cmidrule(lr){2-6}
 & {0} & {1000} & {2000} & {3000} & {4000} \\
\midrule
From I2V & 0.337 & 0.240 & 0.221 & 0.150 & 0.103 \\
Ours     & \textbf{0.110} & \textbf{0.166} & \textbf{0.098} & \textbf{0.103} & \textbf{0.088} \\
\bottomrule
\end{tabular}

\caption{\textbf{Ablation study on base model initialization.}
DreamSim comparison between I2V-based and inpainting-based initialization across training iterations.}
\label{tab:ablation_base_model}
\end{table}

\begin{figure}
\centering
    \includegraphics[width=1.0\columnwidth, trim={1cm 0cm 0cm 0cm}]{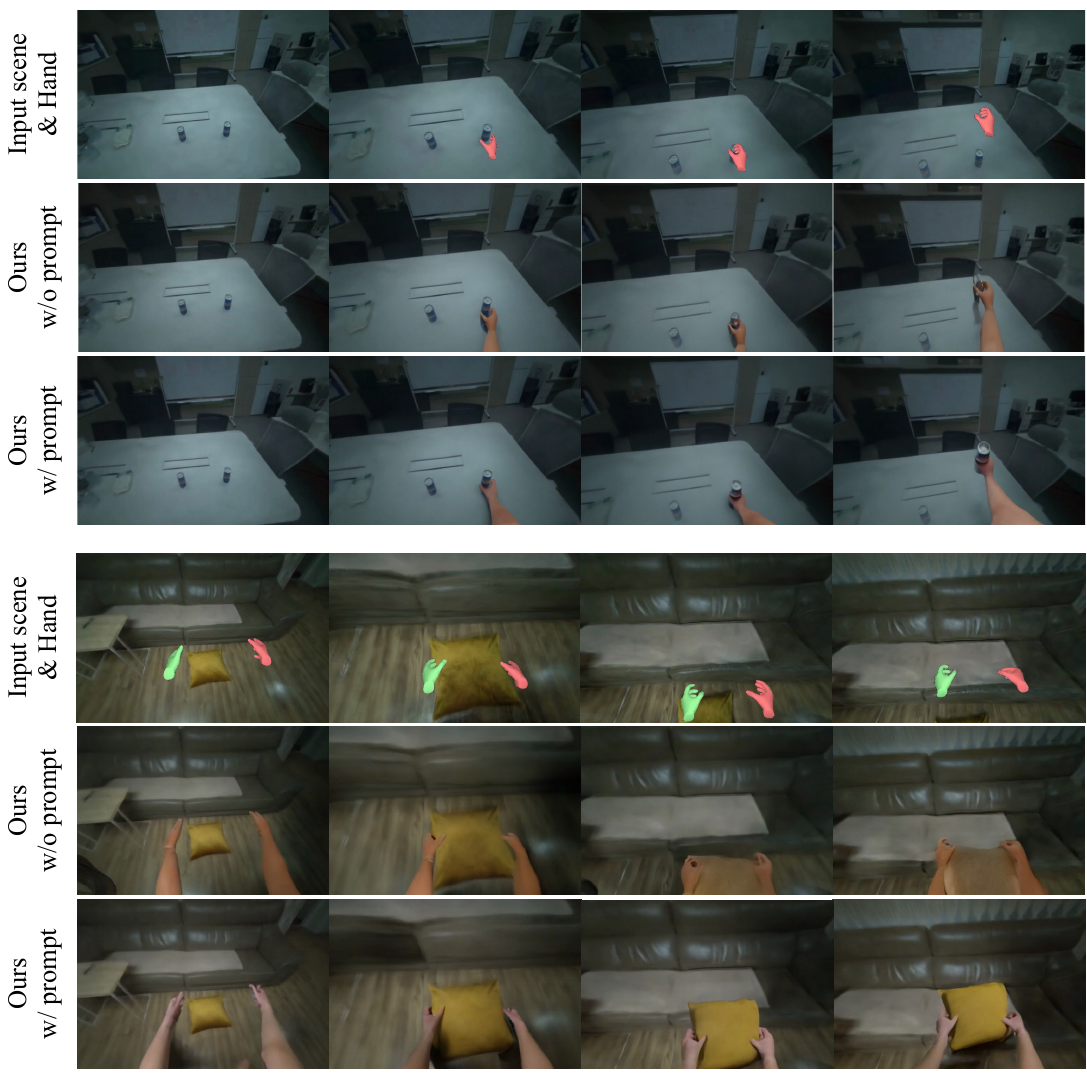}
    \caption{\textbf{Qualitative comparison of DWM outputs for real-world sequences with prompt and without prompt.}}
\label{fig:supp_text_effect}
\end{figure}

\section{Effect of Text Prompts}
\label{sec:text_effect}
Since our model is initialized from a pretrained text-guided video inpainting diffusion model~\cite{VideoX-Fun}, it takes a text prompt as an additional conditioning signal. 
In \cref{fig:supp_text_effect}, we compare results with and without text prompts on real-world interaction sequences. 
Even without textual input, our model generates object motions induced by the given hand manipulation. 
However, the absence of text prompts leads to weaker object consistency and degraded motion accuracy. 
Incorporating text prompts helps stabilize object identity and improves the fidelity of manipulation outcomes, suggesting that textual priors complement our action-driven conditioning by providing high-level semantic guidance.

\begin{figure}
\centering
    \includegraphics[width=1.0\columnwidth, trim={1cm 0cm 0cm 0cm}]{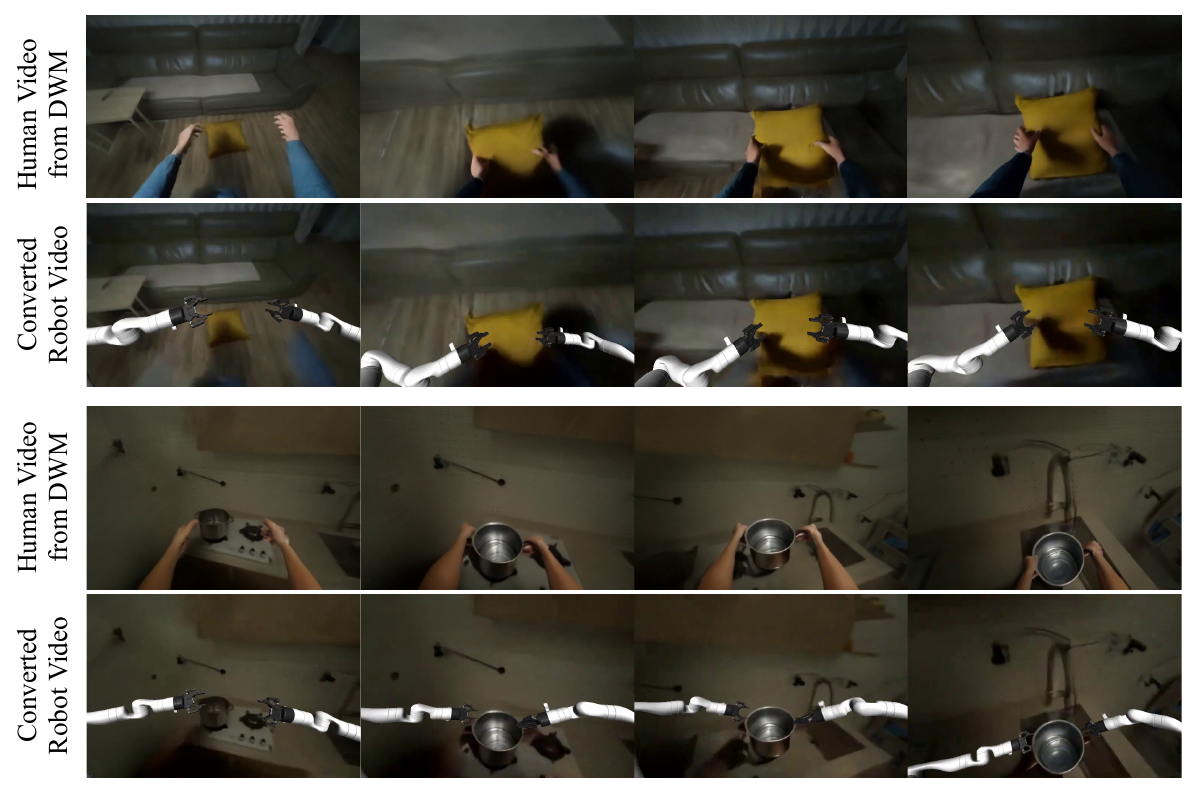}
    \caption{\textbf{Controllable Robot Video Generation.} We can convert the DWM-generated human-scene interaction videos to robot-sene interaction videos following \citet{lepert2025masquerade}.}
\label{fig:supp_robot_video}
\end{figure}
\section{Controllable Robot Video Generation}
Since DWM synthesizes plausible egocentric interaction videos conditioned on hand manipulation trajectories, 
it can be naturally extended to controllable robot video generation by converting human–scene interaction videos into robot–scene interaction videos. 
Specifically, we adopt the approach of \citet{lepert2025masquerade}, which replaces human hands in egocentric videos with a simulated robot arm. 
Their method first masks out the human body from the egocentric view and then renders a robot arm by mapping the human hand pose to a corresponding robot control configuration.
As illustrated in \cref{fig:supp_robot_video}, we apply this pipeline to videos generated by DWM, converting egocentric human manipulation sequences into robot manipulation videos while preserving the underlying interaction dynamics.
This demonstrates that DWM not only enables controllable human-centric video generation, but also serves as a scalable visual simulator for robot manipulation. 
Such robot-centric videos can further support robot policy learning without explicit physics simulators, while also providing an effective source of data augmentation for real-world robot manipulation tasks.~\cite{lepert2025phantomtrainingrobotsrobots}.

\section{Limitations}
\label{sec:limitations}
While our model opens a new direction for simulating embodied dexterous actions in 3D scenes, several limitations remain. 
As discussed in \cref{sec:text_effect}, our framework still relies on text prompts to achieve the best visual quality, since current video diffusion models heavily depend on textual guidance for semantic grounding. 
Distilling such semantic priors into purely action-based control without text supervision remains an important future research direction.
In addition, the model struggles with non-rigid or highly deformable objects, occasionally failing to maintain object rigidity and visual consistency during complex manipulations. 
This limitation likely stems from the limited diversity of deformable-object interactions in our training data and could be alleviated with richer and more varied manipulation datasets.
Moreover, our current pipeline for constructing the real-world dataset with dynamic views lacks scalability. While it is sufficient for use as an evaluation set, scaling it to a size suitable for training remains challenging. Data augmentation may help increase variability, but a truly scalable solution would require automating both acquisition and reconstruction to reduce dependence on manual scene exploration and preprocessing.
Finally, our model does not explicitly reason about 3D structure, depth, or physical contact. 
This restricts its ability to enforce strict physical constraints, which is particularly important for downstream applications such as robotic policy learning. 
Future work includes incorporating depth-aware priors, modeling contact dynamics more explicitly, and using DWM as a differentiable simulator for action optimization and policy learning.

\begin{figure*}
\centering
\includegraphics[width=0.8\linewidth, trim={0cm 0cm 0cm 0cm}]{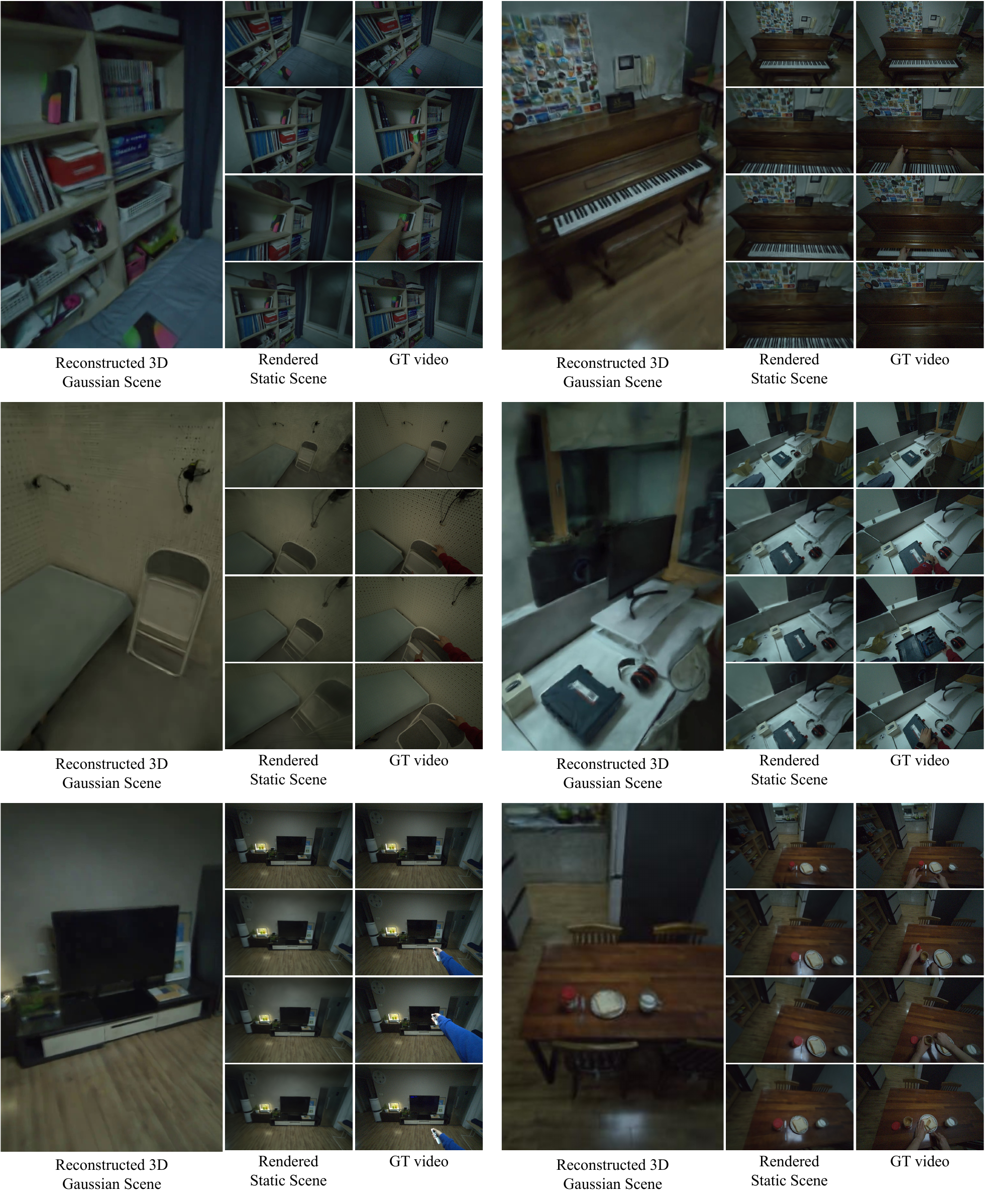}
\caption{\textbf{Examples of paired data of our real-world dataset with dynamic view.} We collect paired sets of static scene video and GT interaction video with our custom data capture protocol using Aria Glasses.}
\label{fig:data_0}
\end{figure*}

\begin{figure*}
\centering
\includegraphics[width=0.9\linewidth, trim={0cm 0cm 0cm 0cm}]{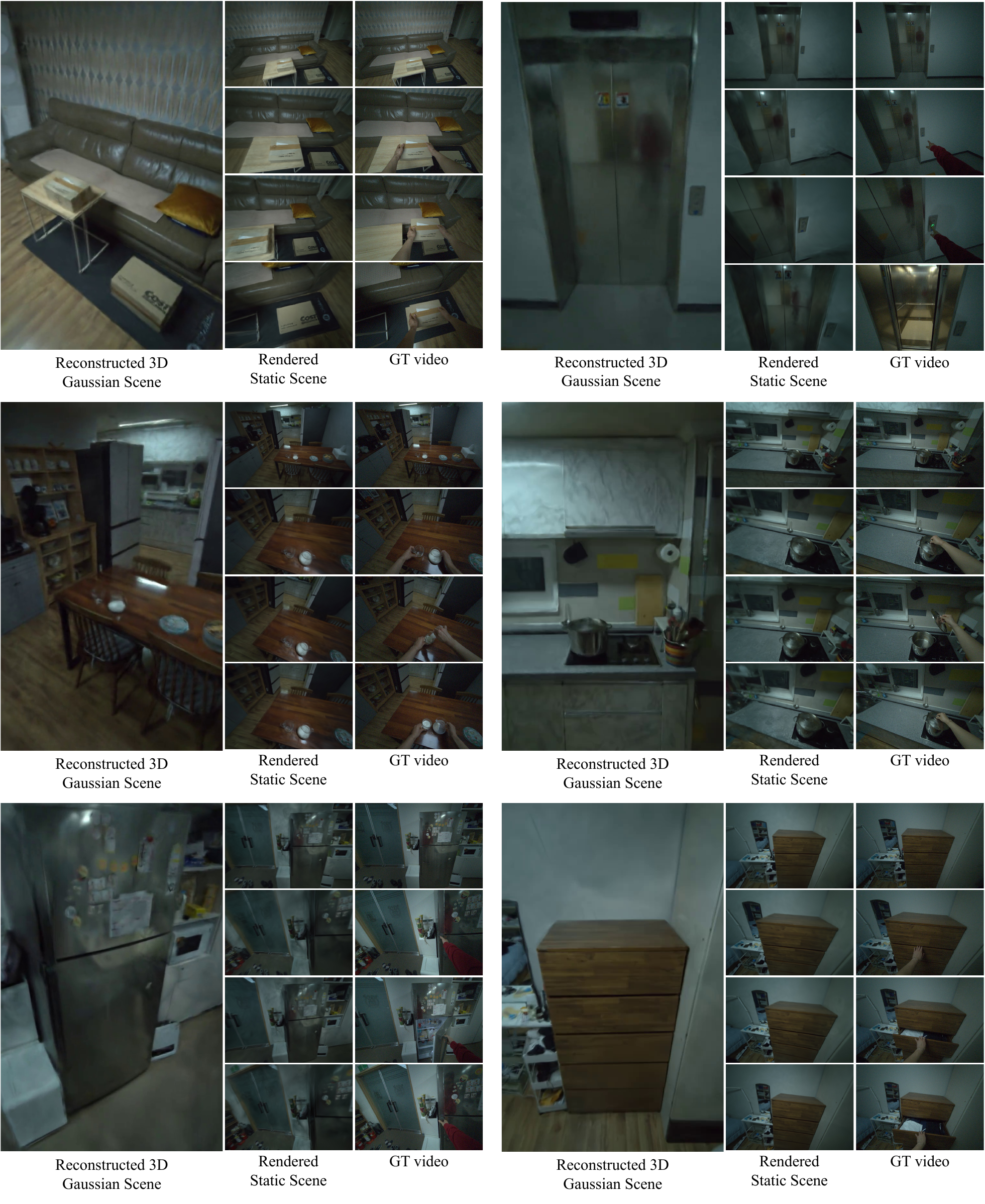}
\caption{\textbf{Examples of paired data of our real-world dataset with dynamic view.} We collect paired sets of static scene video and GT interaction video with our custom data capture protocol using Aria Glasses.}
\label{fig:data_1}
\end{figure*}

\begin{figure*}
\centering
\includegraphics[width=0.9\linewidth, trim={1cm 0cm 0cm 0cm}]{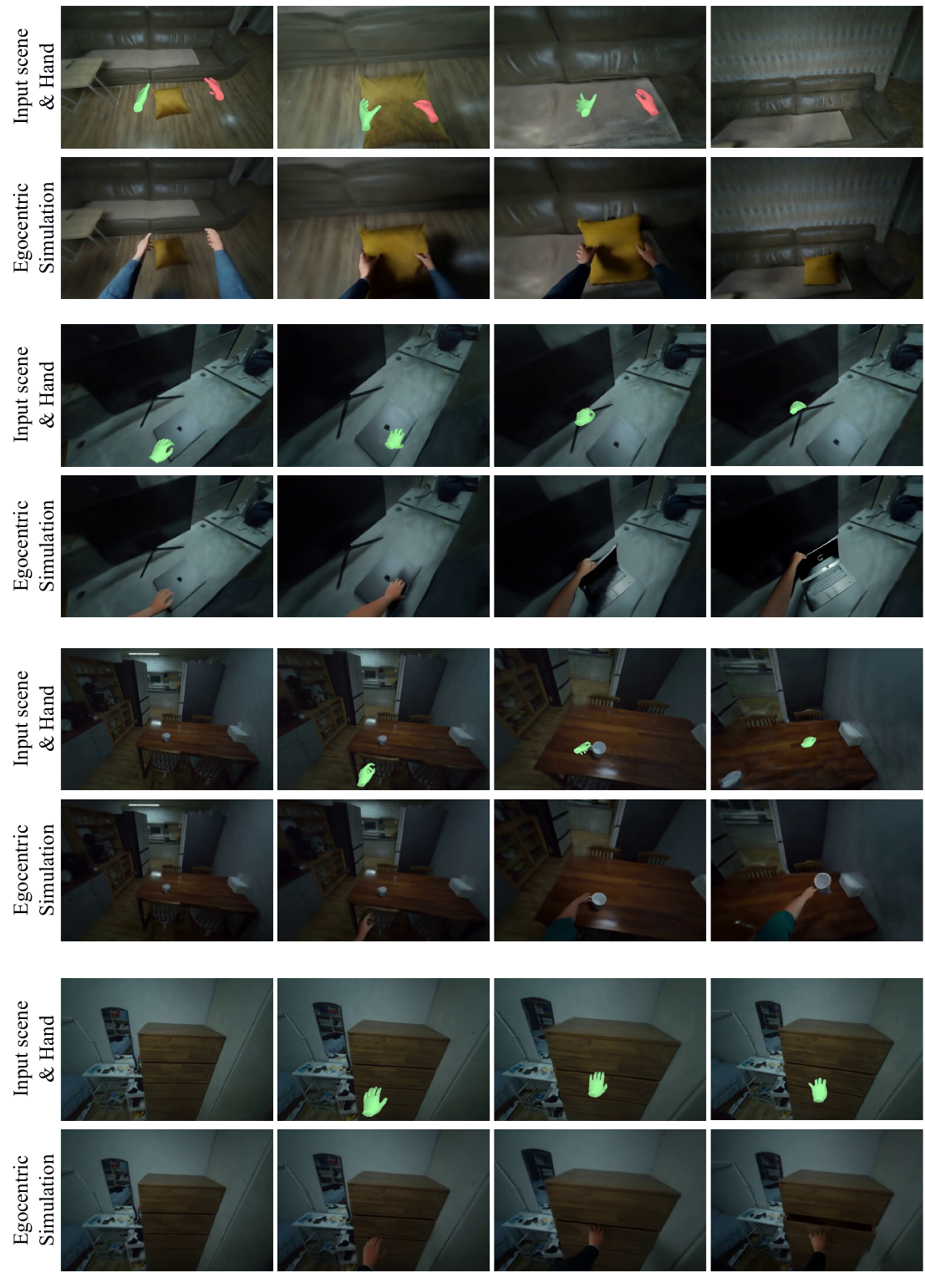}
\caption{\textbf{Additional qualitative results on real-world scene with dynamic view.} We show the input static scene video with hand action in the first row of each pair. The resulting egocentric simulation generated via DWM is demonstrated in the second row.}
\label{fig:extra_4}
\end{figure*}

\begin{figure*}
\centering
\includegraphics[width=0.8\linewidth, trim={1cm 0cm 0cm 0cm}]{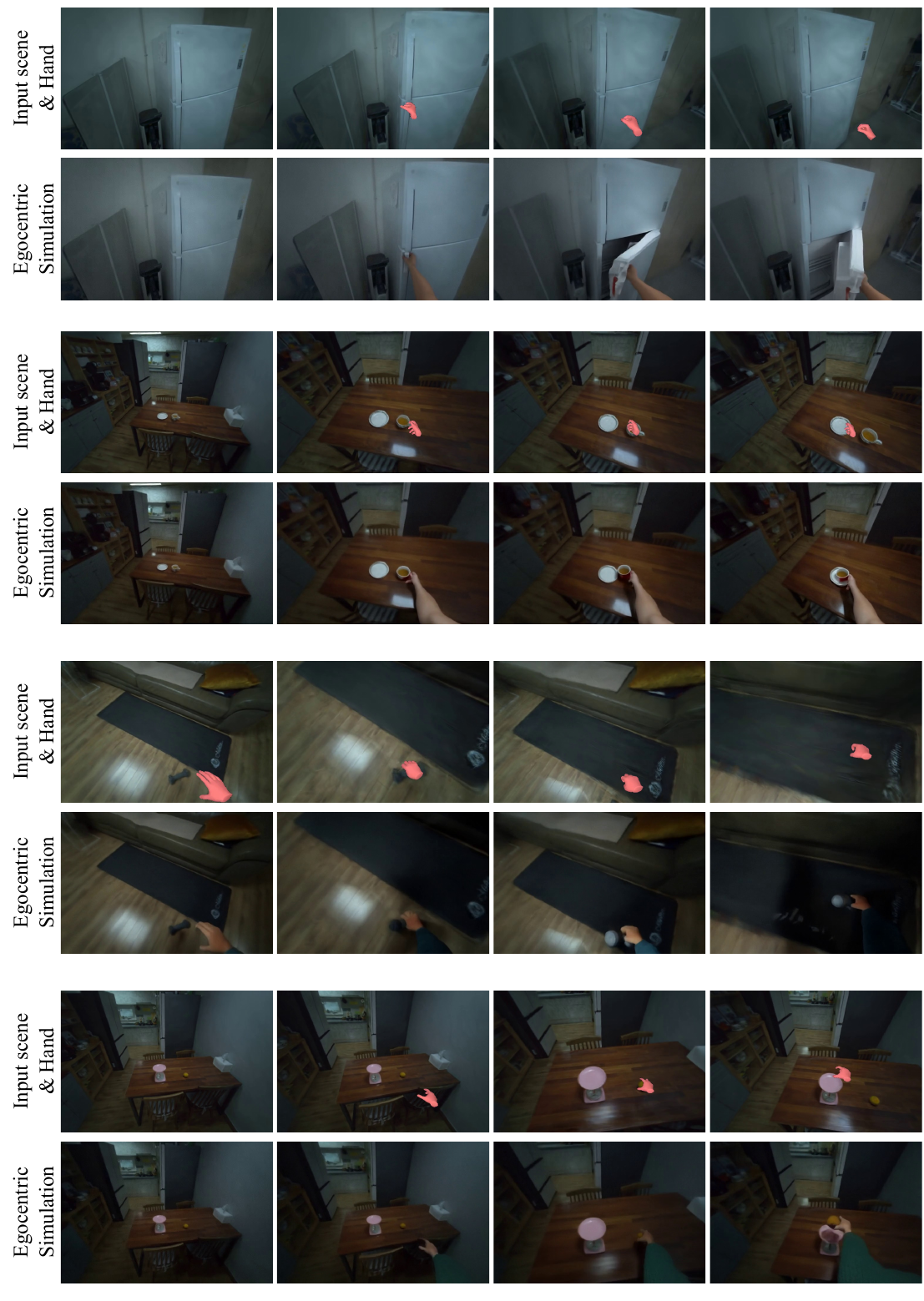}
\caption{\textbf{Additional qualitative results on real-world scene with dynamic view.} We show the input static scene video with hand action in the first row of each pair. The resulting egocentric simulation generated via DWM is demonstrated in the second row.}
\label{fig:extra_5}
\end{figure*}

\begin{figure*}
\centering
\includegraphics[width=0.8\linewidth, trim={1cm 0cm 0cm 0cm}]{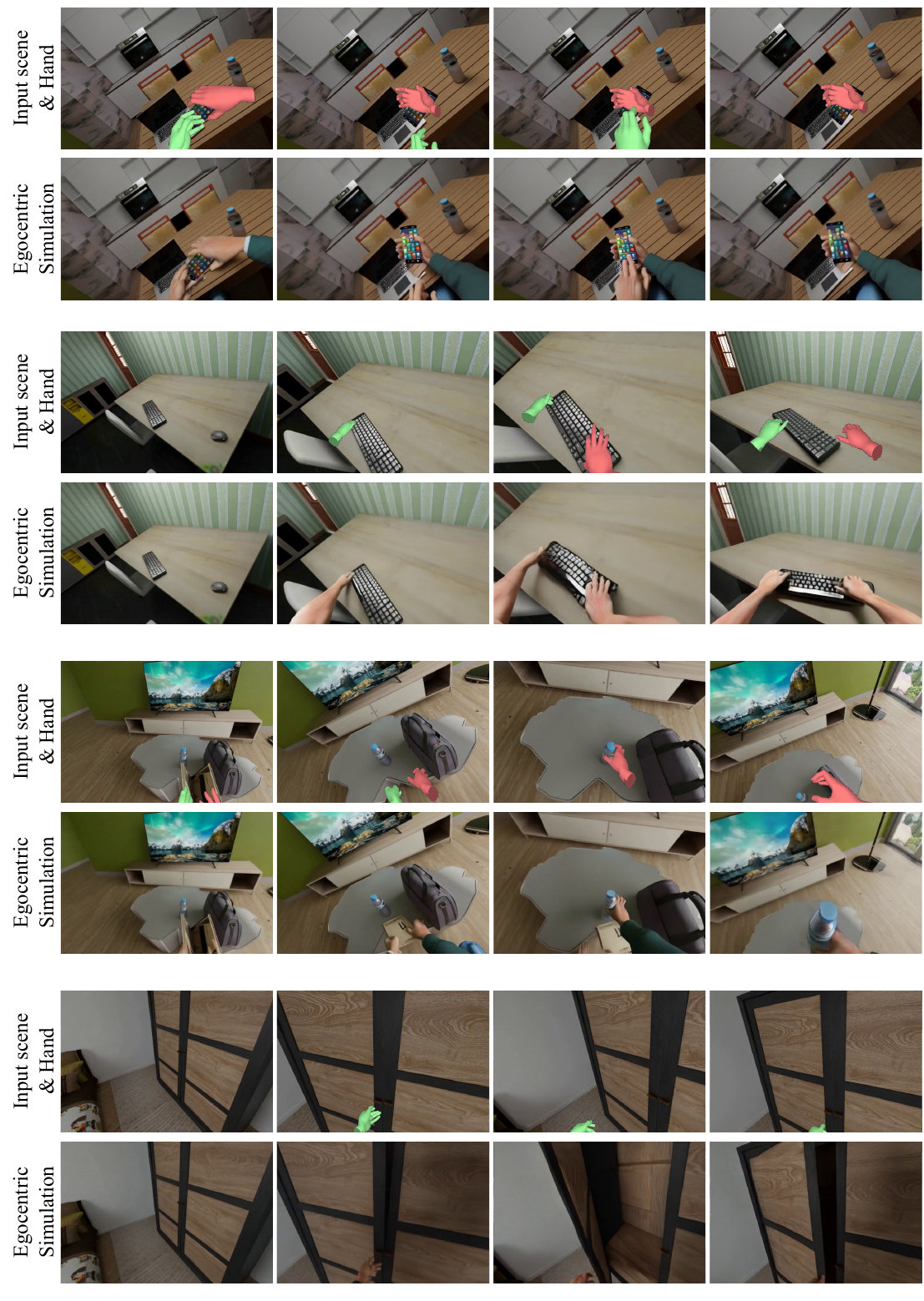}
\caption{\textbf{Additional qualitative results on synthetic scene with dynamic view.} We show the input static scene video with hand action in the first row of each pair. The resulting egocentric simulation generated via DWM is demonstrated in the second row.}
\label{fig:extra_1}
\end{figure*}

\begin{figure*}
\centering
\includegraphics[width=0.8\linewidth, trim={1cm 0cm 0cm 0cm}]{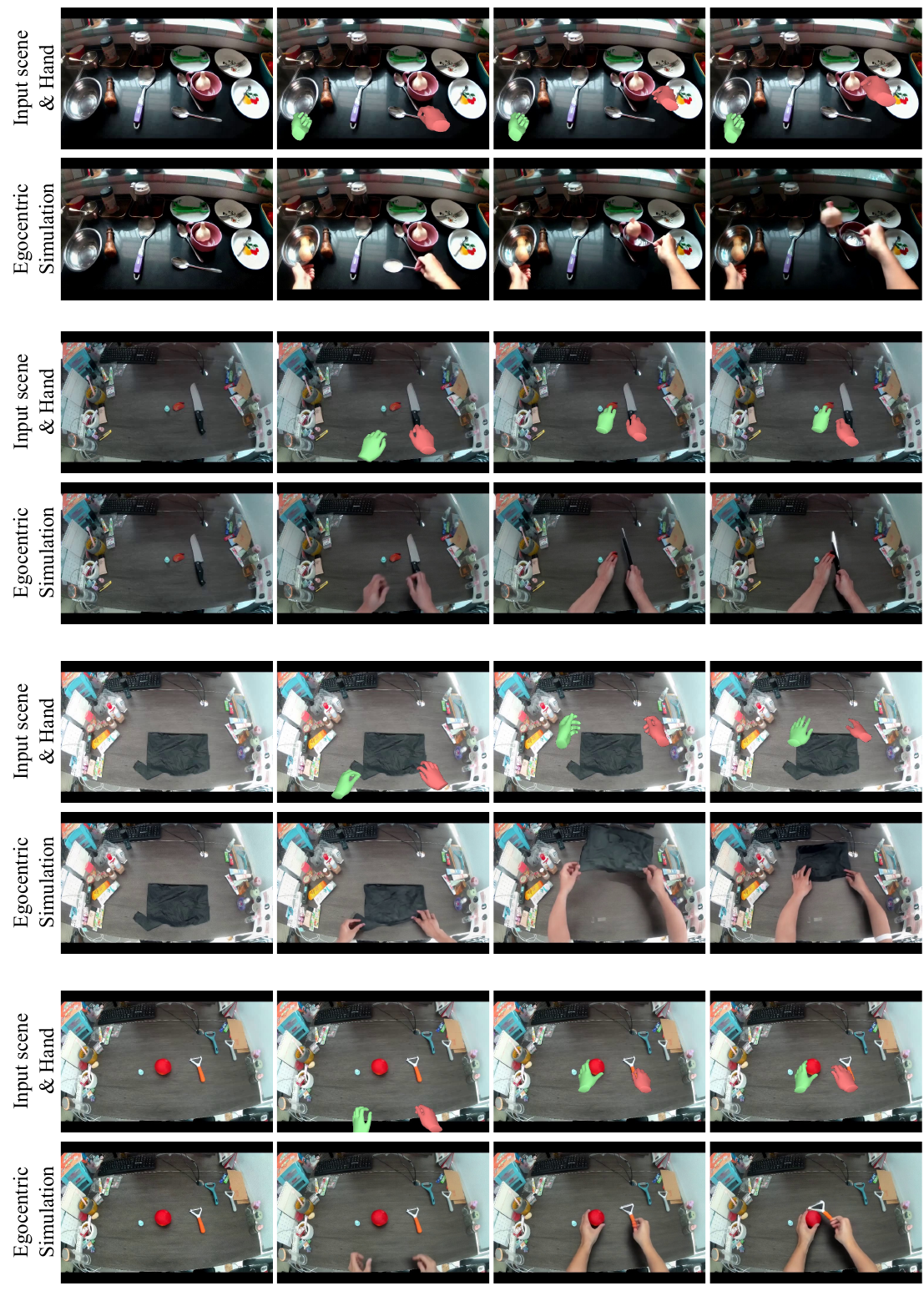}
\caption{\textbf{Additional qualitative results on real-world scene with static view.} We show the input static scene video with hand action in the first row of each pair. The resulting egocentric simulation generated via DWM is demonstrated in the second row.}
\label{fig:extra_2}
\end{figure*}

\begin{figure*}
\centering
\includegraphics[width=0.8\linewidth, trim={1cm 0cm 0cm 0cm}]{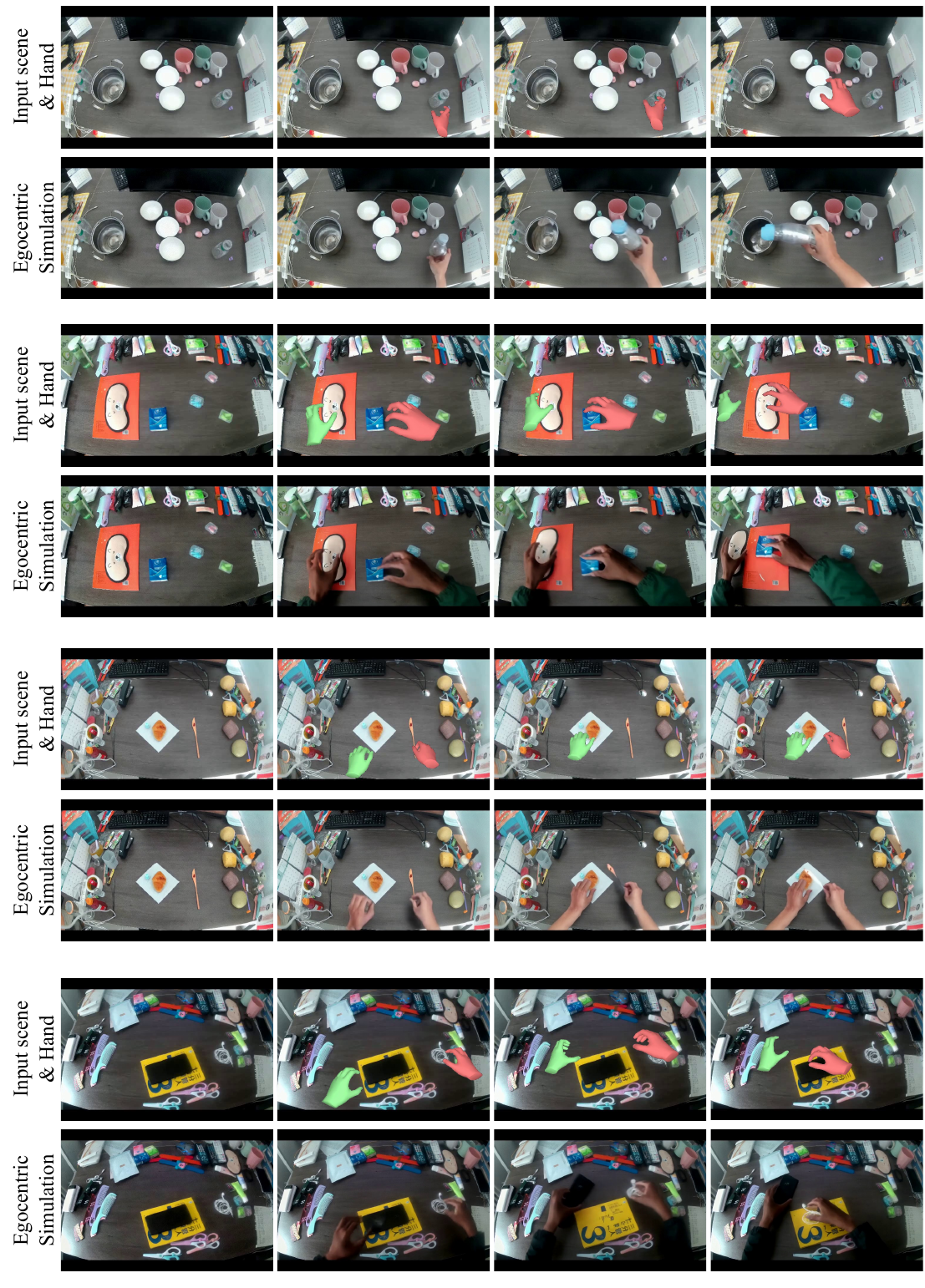}
\caption{\textbf{Additional qualitative results on real-world scene with static view.} We show the input static scene video with hand action in the first row of each pair. The resulting egocentric simulation generated via DWM is demonstrated in the second row.}
\label{fig:extra_3}
\end{figure*}

\fi

\end{document}